\documentclass[10pt,journal,compsoc]{IEEEtran}

\usepackage{cite}
\usepackage{url}
\usepackage{ragged2e}
\usepackage{epsfig}
\usepackage{graphicx}
\usepackage{amsmath,amssymb} 
\usepackage{subfigure}
\usepackage{algorithm}
\usepackage{algorithmicx}
\usepackage{algpseudocode}
\usepackage{graphics}
\usepackage{threeparttable}
\usepackage{color}
\usepackage[normalem]{ulem}
\usepackage{multirow}
\usepackage{float}
\usepackage{amsfonts}
\usepackage{bm}
\usepackage{array}
\usepackage[table]{xcolor}
\usepackage{colortbl}
\usepackage{pifont}
\usepackage{diagbox}
\usepackage{rotating}
\usepackage{booktabs}
\usepackage{overpic}
\usepackage{textcomp}
\usepackage{contour}
\usepackage{enumitem}
\usepackage[pagebackref=false,breaklinks=true,colorlinks=true, bookmarks=false]{hyperref}

\def\ie{\emph{i.e.}}
\def\eg{\emph{e.g.}}
\def\etc{\emph{etc}}
\def\etal{{\em et al.~}}

\definecolor{bblue}{rgb}{0,150,230}
\definecolor{mygray}{gray}{.92}

\newcommand{\figref}[1]{Fig.~\ref{#1}}
\newcommand{\tabref}[1]{Tab.~\ref{#1}}
\newcommand{\eqnref}[1]{(Eq.~\ref{#1})}
\newcommand{\secref}[1]{Sec.~\ref{#1}}

\def\ourmodel{\textit{GCoNet+}}

\graphicspath{{./Imgs/}}
\DeclareGraphicsExtensions{.jpg,.pdf,.png}
\begin{document}
\def\numDeepMethods{11~}

\title{GCoNet+: A Stronger Group Collaborative  Co-Salient Object Detector}

\author{Peng Zheng,
        Huazhu Fu, 
        Deng-Ping~Fan$^{\dagger}$,
        Qi Fan,
        Jie Qin,
        Yu-Wing Tai, \\
        Chi-Keung Tang,
        and Luc Van Gool
\IEEEcompsocitemizethanks{
\IEEEcompsocthanksitem Peng Zheng is with the College of Computer Science and Technology, Nanjing University of Aeronautics and Astronautics, Nanjing, China, with Mohamed bin Zayed University of Artificial Intelligence (MBZUAI), Abu Dhabi, United Arab Emirates, and also with Aalto University.
\IEEEcompsocthanksitem Huazhu Fu is with the Institute of High Performance Computing (IHPC), Agency for Science, Technology and Research (A*STAR), Singapore. 
\IEEEcompsocthanksitem Deng-Ping Fan is with the Computer Vision Lab (CVL), ETH Zurich, Zurich, Switzerland.
\IEEEcompsocthanksitem Qi Fan is with the Hong Kong University of Science and Technology.
\IEEEcompsocthanksitem Jie Qin is with the College of Computer Science and Technology, Nanjing University of Aeronautics and Astronautics, Nanjing, China.
\IEEEcompsocthanksitem Yu-Wing Tai is with Kuaishou Technology and the Hong Kong University of Science and Technology.
\IEEEcompsocthanksitem Chi-Keung Tang is with the Hong Kong University of Science and Technology.
\IEEEcompsocthanksitem Luc Van Gool is with the Computer Vision Lab, ETH Zurich, Zurich, Switzerland, and with KU Leuven, Leuven, Belgium. 
\IEEEcompsocthanksitem A preliminary version of this work has appeared in CVPR 2021~\cite{GCoNet}
\IEEEcompsocthanksitem 
Peng Zheng and Huazhu Fu share equal contributions. 
\IEEEcompsocthanksitem $\dagger$ The major part of this work was done while Peng Zheng was an intern at IIAI mentored by Deng-Ping Fan (Corresponding author: dengpfan@gmail.com).}
}

\markboth{IEEE TRANSACTIONS ON PATTERN ANALYSIS AND MACHINE INTELLIGENCE}%
{Zheng \MakeLowercase{\textit{et al.}}: GCoNet+: A Stronger Group Collaborative Co-Salient Object Detector}

\IEEEtitleabstractindextext{
\begin{abstract}
   \justifying
   In this paper, we present a novel end-to-end group collaborative learning network, termed \textbf{\ourmodel{}}, which can effectively and efficiently (250 fps) identify co-salient objects in natural scenes. The proposed \ourmodel{}~achieves the new state-of-the-art performance for co-salient object detection (CoSOD) through mining consensus representations based on the following two essential criteria: 
   1) \textbf{intra-group compactness} to better formulate the consistency among co-salient objects by capturing their inherent shared attributes using our novel group affinity module (GAM); 
   2) \textbf{inter-group separability} to effectively suppress the influence of noisy objects on the output by introducing our new group collaborating module (GCM) conditioning on the inconsistent consensus. To further improve the accuracy, we design a series of simple yet effective components as follows: 
   i) a recurrent auxiliary classification module (RACM) promoting model learning at the semantic level;
   ii) a confidence enhancement module (CEM) assisting the model in improving the quality of the final predictions; and
   iii) a group-based symmetric triplet (GST) loss guiding the model to learn more discriminative features.
   Extensive experiments on three challenging benchmarks, \ie{}, CoCA, CoSOD3k, and CoSal2015, demonstrate that our \ourmodel{}~outperforms the existing 12 cutting-edge models. 
   Code has been released at \url{https://github.com/ZhengPeng7/GCoNet_plus}.
\end{abstract}

\begin{IEEEkeywords}
Co-saliency, CoSOD, Group Collaborative Learning, Deep Learning.
\end{IEEEkeywords}}

\maketitle
\IEEEdisplaynontitleabstractindextext
\IEEEpeerreviewmaketitle


\IEEEraisesectionheading{\section{Introduction}\label{sec:introduction}}
\IEEEPARstart{C}{o-Salient} object detection (CoSOD) aims at detecting the most common salient objects among a group of given relevant images. 
Compared with the standard salient object detection (SOD) task, CoSOD is more challenging and requires distinguishing co-occurring objects across different images where others act as distractors. To this end, intra-class compactness and inter-class separability are two important cues and should be learned simultaneously. 
With the increasing accuracy and efficiency achieved by the latest CoSOD methods, CoSOD is not only used as a pre-processing component for other vision tasks~\cite{wang2016higher,hsu2019deepco3,zeng2019joint,jerripothula2018efficient,wang2019no} but also employed in many practical applications~\cite{GCoNet,CoSal2015,SOD_review3}.

\begin{figure}[t!]
	\centering
	\includegraphics[width=.9\linewidth]{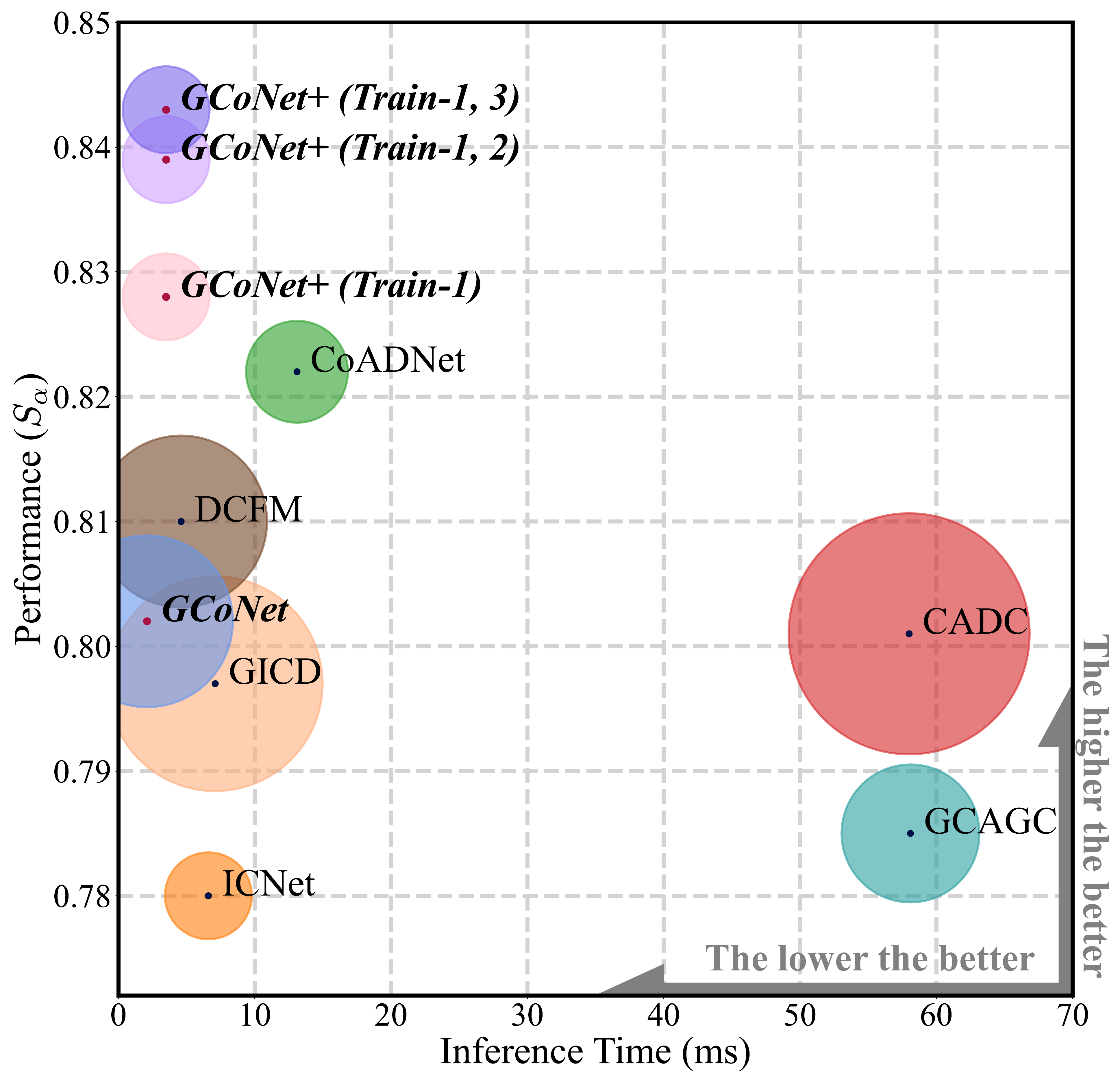}
	\vspace{-10pt}
	\caption{\textbf{Comparisons of seven representative CoSOD approaches and ours on the CoSOD3k dataset~\cite{CoSOD3k}}. We conduct the comparison of existing representative deep-learning-based CoSOD approaches in terms of both speed (the horizontal axis) and accuracy 
	(the vertical axis). Smaller bubbles mean lighter models.
	Our \ourmodel{}~outperforms these models in terms of both efficiency and effectiveness. The ``Train-1, 2, and 3'' represents the DUTS\_class, COCO-9k, and COCO-SEG datasets, respectively (see \tabref{table:ablation_results} for more related details). All the models are tested with batch size 2 on an A100-80G. Our benchmark for inference speed can be found at \url{https://github.com/ZhengPeng7/CoSOD_fps_collection}.
	}\label{fig:Acc-speed}
\end{figure}

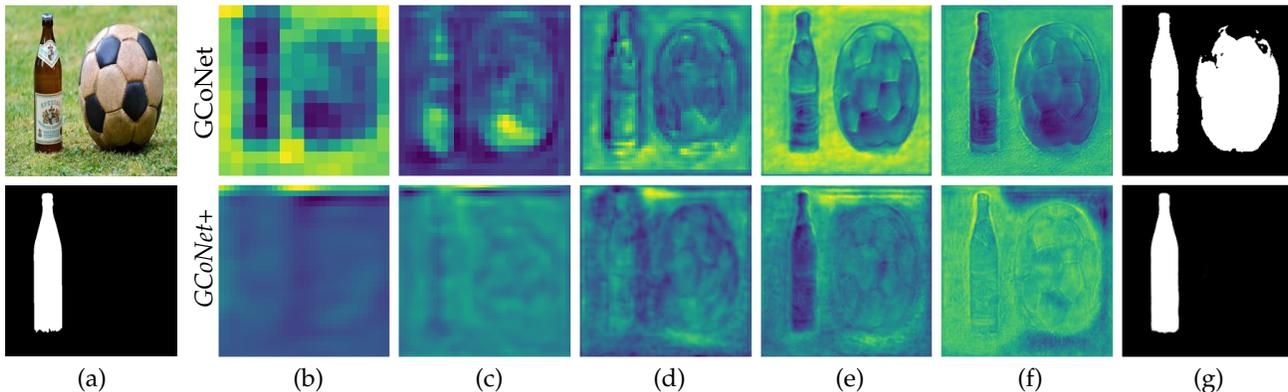
\begin{figure*}[t!]
	\centering
    \begin{overpic}[width=.95\textwidth]{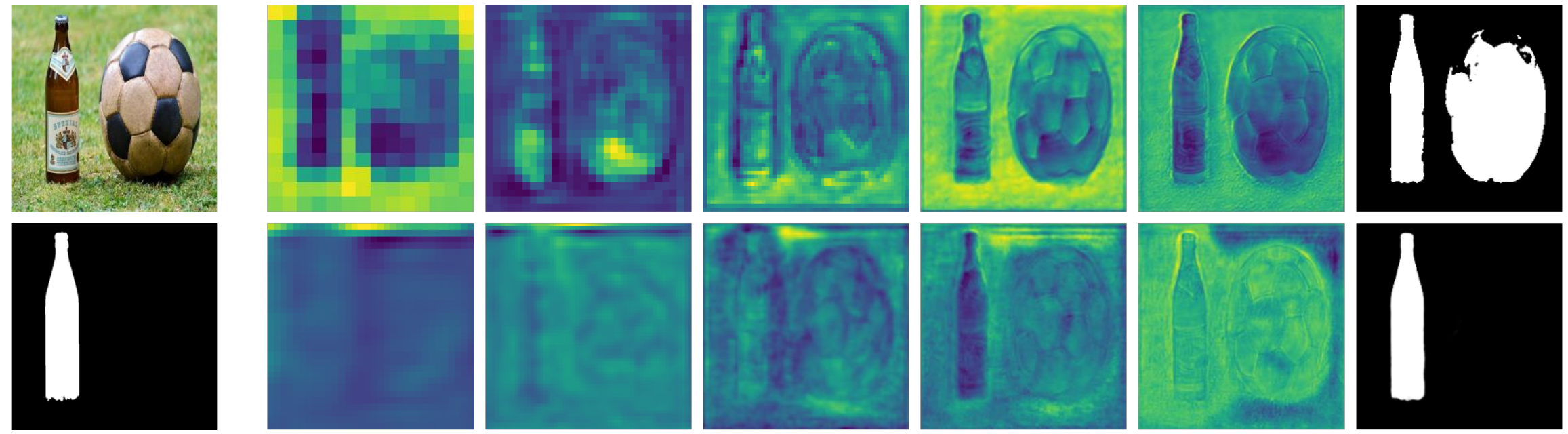} 
        \put(15.2,17){\rotatebox{90}{GCoNet}}
        \put(15.2,4){\rotatebox{90}{\ourmodel{}}}
        \put(6.2,-2){(a)}
        \put(22.8,-2){(b)}
        \put(36.8,-2){(c)}
        \put(50.5,-2){(d)}
        \put(64.5,-2){(e)}
        \put(78.5,-2){(f)}
        \put(92,-2){(g)}
    \end{overpic}
    \caption{\textbf{Visualizations of feature maps.} 
    (a) Source image and ground truth. 
    (b-f) Feature maps on different levels of the decoder from our GCoNet~\cite{GCoNet} and \ourmodel{}~(Train-1 in \tabref{table:ablation_results}), captured from high to low levels. The feature maps shown in (b) have the lowest resolution. As (b) shows, our \ourmodel{}~gives a more global response and does not make a specific prediction in the very early stages, where the quality of feature maps is inadequate for producing precise results.
    (g) Prediction of co-saliency maps. Compared with GCoNet, 
    \ourmodel{}~obtained a more global response on the objects and its surrounding.}
    \label{fig:fpn_features}
\end{figure*}

Existing works attempt to facilitate the consistency among given images to solve the CoSOD task \textit{within an image group} by leveraging semantic connections~\cite{zhang2016co,MetricCoSOD,hsu2018unsupervised} or varied shared cues~\cite{li2011co,CBCS,cao2014self}. In~\cite{CoSOD3k,GICD}, the proposed models jointly optimize a unified network for generating saliency maps and co-saliency information.
Despite the improvement brought by these methods, most existing models only depend on the consistent feature representations in an individual group~\cite{GICD,GWD,CoADNet,GLNet,ICNet,SRG}, which may introduce the following limitations. First, images from the same group can only provide positive relations instead of both positive and negative relations between different objects. Training models with only positive samples from a single group may lead to overfitting and ambiguous results for outlier images. Besides, there are typically a limited number of images in one group (20 to 40 images for most groups on existing CoSOD datasets). Therefore, information learned from a single group is usually insufficient for a discriminative representation. Finally, individual image groups may not be easy to mine semantic cues, which are vital in distinguishing noisy objects during testing in complex real-world scenes. Due to the complexity of image context in real scenarios, a module designed for common information mining is in high demand. Apart from these, when supervised with the Binary Cross Entropy (BCE) loss, pixel values of generated saliency maps tend to get closer to 0.5 instead of 0 or 1. Suffering from the uncertainty, these maps are difficult to directly apply in realistic applications.

To overcome the above restrictions, we propose a new group collaborative learning network (GCoNet), which establishes semantic consensus within the same group and distinction among different image groups. 
Our GCoNet includes three basic modules: Group Affinity Module (GAM), Group Collaborating Module (GCM), and Auxiliary Classification Module (ACM), which simultaneously guide our GCoNet to learn the
\textbf{inter-group separability} and \textbf{intra-group compactness} in a better way. Specifically, the GAM enables the model to learn the consensus feature in the same image group, while the GCM discriminates target attributes among different groups, 
thus making the network trainable on existing rich SOD datasets\footnote{There are about 60k SOD images publicly available, which is about 10 times larger than existing CoSOD datasets. This means the insufficient training data issue in CoSOD may be partially alleviated in the proposed framework.}. To learn a better embedding space, we utilize the ACM on each image to improve the feature representation at a global semantic level.

We have improved our original GCoNet by providing a more precise explanation of the existing contributions, \emph{i.e.}, 
a concise network for CoSOD, 
three additional components that can improve the ability to learn consensus and difference, 
discussions on the shortcomings of existing training sets and the corresponding solution.

In summary, we have extended our GCoNet significantly to \ourmodel{} with major differences as follows:

\begin{itemize}\setlength\itemsep{0.3em}
    \item 
    \textbf{Novel Approaches.}
    We propose three new components which improve the performance and robustness of our \ourmodel{}, \ie{}, Confidence Enhancement Module (\textbf{CEM}), Group-based Symmetric Triplet (\textbf{GST}) Loss, and Recurrent Auxiliary Classification Module (\textbf{RACM}) to deal with the existing weaknesses of our GCoNet.
    1) Confidence Enhancement Module (\textbf{CEM}): To make output maps less uncertain, we employ the differentiable binarization and a hybrid saliency loss in our confidence enhancement module, which brings maps of higher quality and further improves the overall performance.
    2) Group-based Symmetric Triplet (\textbf{GST}) Loss: We are one of the first ones to apply metric learning to deep-based CoSOD models, which makes the learned features of different groups more discriminative in a metric learning way.
    3) Recurrent Auxiliary Classification Module (\textbf{RACM}): To better represent the auxiliary classification feature, we extend the original auxiliary classification module to its recurrent version, which focuses more precisely on the pixels of target objects.
    Besides, we improve the GCoNet~\cite{GCoNet} into a more lightweight and powerful network as our baseline.
    These three components and the new baseline network are organically combined and achieved great performance on all existing datasets and realistic applications in our experiments.

    \item
    \textbf{Experiments.}
    Although the development of Co-Salient Object Detection is rapid, there are three usually used datasets for training, \ie{}, DUTS\_class, COCO-9k, and COCO-SEG, while there is no standard for choosing the training sets for this task. Different from existing works where the used training sets are not the same, we conduct more comprehensive experiments with all different combinations of these three training sets for fair experimental comparisons.
    With the combinations of the newly proposed components in this paper, as mentioned above in novel approaches, we obtain $\sim$3.2\% relative improvement on the $E_{\xi}^\text{max}$~\cite{Emeasure} and $S_\alpha$~\cite{Smeasure} compared with our GCoNet~\cite{GCoNet} with the same training set, achieving the state-of-the-art performance to this day among all publicly available CoSOD models~\cite{CoSOD3k}\footnote{Public leaderboard of CoSOD models: \url{https://paperswithcode.com/task/co-saliency-detection}}.

    \item
    \textbf{New Insights.}
    Based on the obtained experimental results, we found the potential problems of the existing CoSOD training sets and provided the corresponding analyses on how to improve them in the future.

\end{itemize}


\section{Related Work}
\label{sec:Related Work}
\subsection{Salient Object Detection}
Among traditional salient object detection (SOD) methods, hand-crafted features play the most important role in detection~\cite{Cheng2011GlobalCB,SOD_trad1,SOD_trad2,SOD_trad3}. In the early years of deep learning, features are extracted from image patches, object proposals~\cite{SOD_op1,SOD_op2,SOD_op3}, or super-pixels~\cite{SOD_patch1,SOD_patch2,SOD_patch3,SOD_patch4}. Although these methods have made some progress, they are time-consuming for extracting the target regions and their features. With the success of fully convolutional networks~\cite{FCN} in segmentation tasks, recent SOD researches mainly focus on the models which make pixel-wise predictions. More details and a summary can be found in recent review works~\cite{SOD_review3,SOD_review1,SOD_review2}, where the latest work~\cite{SOD_review3} provides the most comprehensive benchmark and analysis on the performance, robustness, and generalization of existing SOD models on a variety of challenging SOD datasets, which also presents a constructive discussion for SOD open issues and future research directions. In~\cite{PoolNet+}, the network architectures in SOD methods are divided into five categories, \ie{}, single-steam, multi-stream, side-fusion, U-shape, and multi-branch. Among these architectures, the U-shape is the most widely used one, especially the base structure of FPN~\cite{FPN} and U-Net~\cite{UNet}. Multi-stage supervision is employed at the early stages by aggregating features from different stages of these U-shape networks to make the output features more robust and stable~\cite{EGNet,GCoNet,GICD}. In~\cite{SOD_att1,SOD_att2,SOD_att3,SOD_att4}, the attention mechanism and the related modules are designed for improvement. Besides, external information is introduced as extra guidance on its training processes, such as edge~\cite{EGNet} and boundary~\cite{BASNet}.

In the binary segmentation tasks (\eg{}, Salient Object Detection~\cite{GCoNet,BASNet,SOD_att1}, Optical Character Recognition~\cite{DB,OCR_1,OCR_2}), ground truths are the binary maps of target objects. However, the predicted maps are not fully binary ones due to pixel-level loss (\ie{}, mean-square error loss, binary cross-entropy loss). In many practical applications, maps with much uncertainty are unsuitable for programs to make decisions~\cite{u2net_ext}. In that case, some up-to-date methods are proposed for improving the quality of binary maps. In~\cite{integrity}, specific components are designed to enhance the integrity of objects. In~\cite{BASNet}, hybrid losses are also employed to make models focus on more attributes beyond pixel-level errors.

\subsection{Image Co-Segmentation}
Image co-segmentation, a fundamental and active computer vision task, segments the common objects from a group of images.  This has been widely adopted in many related areas, such as co-salient object detection~\cite{GWD,GCoNet,CoSOD3k,arxiv_UFO}, few-shot learning~\cite{coseg_fss1,coseg_fss2}, semantic segmentation~\cite{coseg_ss1,coseg_ss2},~\etc{}. Many existing methods in co-segmentation employ the Siamese network to find the common feature of the input image pair~\cite{coseg1,coseg2}. Based on the comparison within the image pair, Chang~\etal{}~\cite{coseg_rw1} and Rother~\etal{}~\cite{coseg_rw2} used saliency and color histograms, respectively to guide a more precise comparison of the visual features. With the development of deep learning methods, co-segmentation models tend to use implicit semantic features to find common objects. From the aspect of the model, Wei~\etal{}~\cite{GWD} and Fan~\etal{}~\cite{CoSOD3k} embedded the co-attention in their network to generate the group consensus, Chen~\etal{}~\cite{coseg_attCha} leveraged channel attentions for better object co-segmentation, and LSTM~\cite{lstm} is employed by Zhang~\etal{}~\cite{coseg_lstm1} and Li~\etal{}~\cite{coseg_lstm2} to exchange the information between two images and enhance the group representations. From the aspect of the training strategy, Wang~\etal{}~\cite{coseg_weakly} explored the saliency-guided iterative refinement on result maps with weakly supervised strategy, and Hsu~\etal{}~\cite{hsu2018co} made use of the intra-image object discrepancy and inter-image figure-ground separation to achieve image co-segmentation in an unsupervised manner.

\subsection{Co-Salient Object Detection}
The SOD task~\cite{SOD1,BASNet,SOD2,SOD3} aims at segmenting salient objects separately in a single image, while CoSOD targets finding common salient objects across a group of semantic-related images. Previous CoSOD methods mainly aimed at mining intra-group cues to segment the co-salient objects. For example, early CoSOD approaches used to explore the correspondence among a group of relevant images based on handcrafted cues. With computational fractions (\eg{}, superpixels~\cite{SLIC}) segmented from each image, these methods establish the correspondence model and discover the common regions by employing a ranking scheme, clustering guidance, or translation alignment~\cite{CoSOD_trad_align}. Metric learning~\cite{MetricCoSOD, MetricCoSOD1}, statistics of histograms and contrasts~\cite{Cheng2011GlobalCB}, and pairwise similarity ranking are also applied to formulate better semantic attributes for further computation.

In the deep learning era, many end-to-end deep CoSOD models have been proposed. The authors of~\cite{GWD,MetricCoSOD} attempt to discover the common objects by learning the consensus in a single group. With the development of 
the upstream deep learning methods, existing methods~\cite{CoSformer,GCoNet,CoADNet,GICD,ren2022adaptive} build their models with powerful CNN models (\eg{}, ResNet~\cite{ResNet}, VGGNet~\cite{VGG}, and Inception~\cite{Inceptionv2v3}) or even Transformer models (\eg{}, ViT~\cite{ViT} and PVT~\cite{PVTv1,PVTv2}), which help achieve SOTA performances.
Besides most existing works designing their models by full supervision, weakly-supervised strategies (\eg{}, GWSCoSal~\cite{qian2022co}, FASS~\cite{FASS}, SP-MIL~\cite{SP_MIL}, CODW~\cite{CoSal2015}, and GONet~\cite{GONet}) achieve acceptable results as well.

\subsection{Intra- and Inter-image Consistency Learning}

With the rapid development of deep learning, deep models achieved great performance in exploring intra- and inter-image consistency, such as graph convolutional networks (GCN)~\cite{jiang2019unified,jiang2019multiple,GCAGC}, co-attention~\cite{CoSOD3k}, co-clustering~\cite{yao2017revisiting}, recurrent units~\cite{RCAN}, correlation techniques~\cite{ICNet}, self-learning methods~\cite{zhang2016co}, and quality measuring~\cite{jerripothula2018quality}. 

Among the implementations of intra-image consistency learning, co-attention is one of the most widely used components for exploring the consensus for segmentation on similar images since it was first explored in~\cite{lu2019see}. Furthermore, many follow-up works~\cite{wang2021exploring,lu2021segmenting,wang2019zero} then dive deeper via more information and better methods, including pixel contrast, relational data, and graph network. These works show great effectiveness and have brought much improvement to the research in related areas.

Besides, intra- and inter-image consistency also show its effectiveness in other research areas, such as object detection~\cite{intra_inter1,intra_inter2}, semantic segmentation~\cite{intra_inter3}, and salient object detection~\cite{intra_inter4}, especially for establishing the relations between objects to obtain better semantic features of different categories on weakly supervised learning.

In previous CoSOD methods, intra-group consistency has been  studied in detail~\cite{GCoNet,GICD,GWD,CoSal2015}.In contrast, less attention has been paid to the inter-group, which, however, contributes significantly to guiding the model to learn more discriminative and general features for each class. In~\cite{GICD}, a jigsaw training strategy is used to introduce images from other groups to implicitly facilitate group training. In~\cite{CoSal2015}, multiple groups of images are fed into their model to learn the intra-image contrast. Without a more advanced and explicit design for learning inter-group information, their models still mainly target intra-group information. Our approach varies a lot from existing models in exploring inter-group relations. We try to learn discriminative features semantically, explicitly, and precisely at a group level.


\section{Methodology}
We introduce our \ourmodel{}~for the CoSOD task. 
The overview of the architecture is presented in \secref{sec:overview}. Then, we sequentially introduce the proposed basic modules: group affinity module (GAM), 
group collaborating module (GCM), 
confidence enhancement module (CEM), 
group-based symmetric triplet (GST) loss, 
and the recurrent auxiliary classification module (RACM).

\subsection{Overview}\label{sec:overview}
The basic framework of the proposed \ourmodel{}~is based on our GCoNet~\cite{GCoNet}, which is one of the latest state-of-the-art methods. Unlike existing CoSOD models~\cite{CoSOD3k,CoADNet,GICD,ICNet} that only exploit the common information inside a single class group, \ourmodel{}~exploits both the internal and external relationship between different groups in a siamese style.

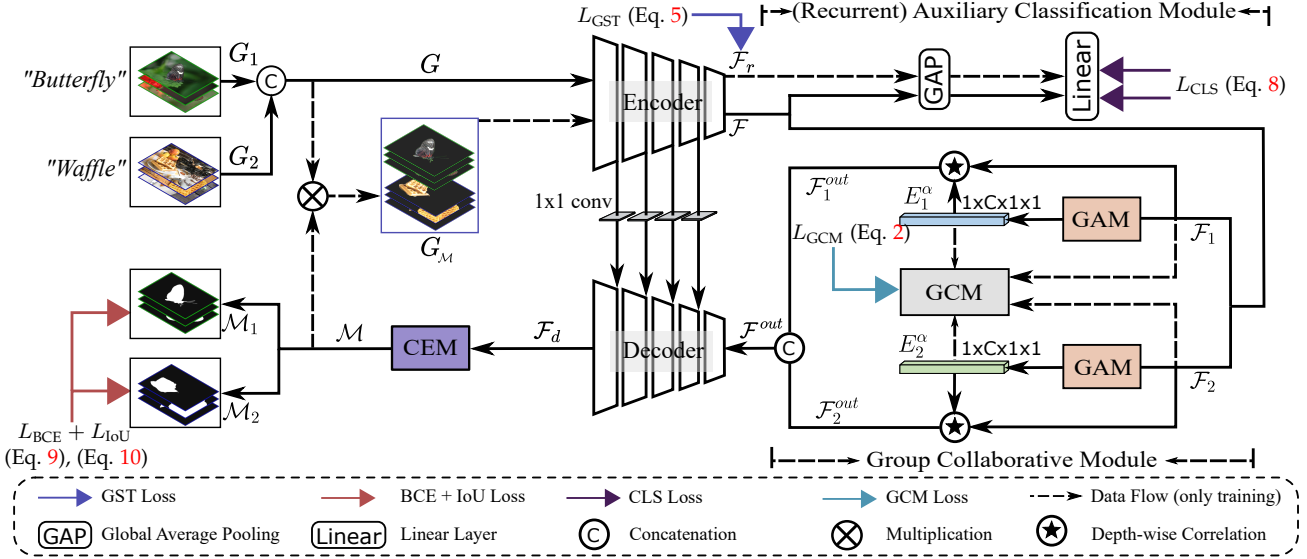
\begin{figure*}[t!]
  \centering
    \begin{overpic}[width=.95\linewidth]{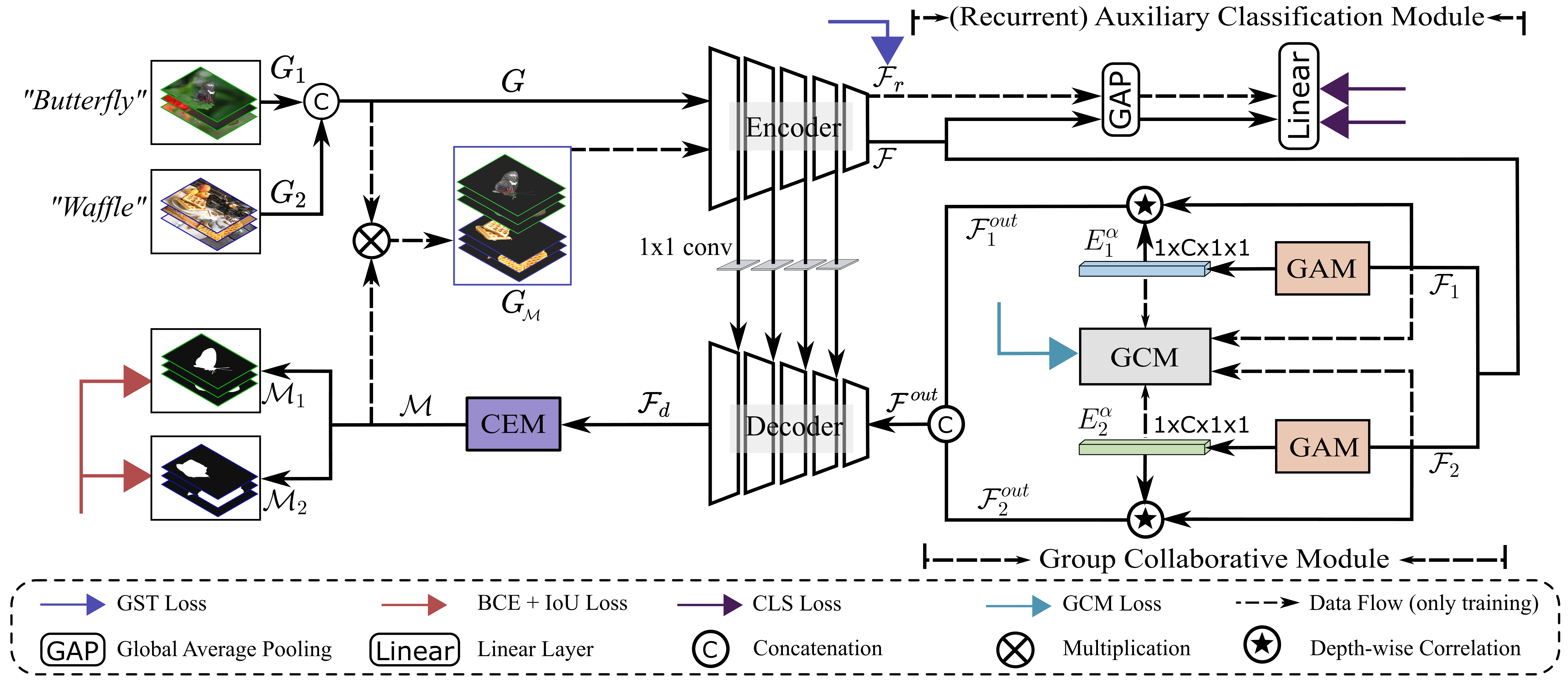}
        \put(1,9.5){\footnotesize {${L}_{\text{BCE}} + {L}_{\text{IoU}}$}}
        \put(0,7.5){\footnotesize {\eqnref{eqn:loss_BCE}, \eqnref{eqn:loss_IoU}}}
        \put(60.5,24.8){\footnotesize {${L}_{\text{GCM}}$ \eqnref{eqn:GCM}}}
        \put(44,41.5){\footnotesize {${L}_{\text{GST}}$ \eqnref{eqn:loss_GST}}}
        \put(90,36.2){\footnotesize {${L}_{\text{CLS}}$ \eqnref{eqn:loss_cls}}}
    \end{overpic}
   \caption{\textbf{Pipeline of the proposed Group Collaborative Learning Network plus (\ourmodel{}).} Input images are obtained from two groups and fed into an encoder. Then we employ the GCoM (Group Collaborative Module), where intra-group collaborative learning is conducted for each group by the group affinity module (GAM), and the inter-group collaborative learning is conducted via the group collaborating module (GCM). The original images and RoIs masked by the output are given to the encoder to do an auxiliary classification to make the features of different classes more discriminative to each other. The decoder output is put through the confidence enhancement module (CEM) to make the final result more binarized and easy to use. Furthermore, the RoIs of two groups obtained by the multiplication of original images and predicted saliency maps are measured with a triplet loss to enlarge the distance between inter-group features and reduce the distance between intra-group features.
   }\label{fig:Arch}
\end{figure*}

The flowchart of \ourmodel{}~is illustrated in \figref{fig:Arch}. First, our model simultaneously takes two groups of raw images ${G_1, G_2}$ as input. With the concatenated image groups ($\copyright$), our encoder extracts the feature maps $\mathcal{F}$, which is then fed to the auxiliary classification module (ACM) for classification and to our group collaborative module (GCoM) for further processing. In GCoM, $\mathcal{F}$ is split into two parts by their classes, \ie{}, $\mathcal{F}_1=\{F_{1,n}\}_{n=1}^N, \mathcal{F}_2=\{F_{2,n}\}_{n=1}^N \in \mathbb{R}^{N \times C \times H \times W}$, where C denotes the channel number, $H \times W$ is the spatial size, and $N$ denotes the group size. These two features are separately given to the group affinity module (GAM), where all the single-image features are combined to distill the consensus features ${\textit{E}_1^{\alpha}} \in \mathbb{R}^{1 \times C \times 1 \times 1}$. Meanwhile, a group collaborating module (GCM) is applied to obtain a more discriminative representation of target attributes between different image groups. The output features ${\mathcal{F}_1^{out}, \mathcal{F}_2^{out}}$ of GCoM are concatenated to be fed to our decoder. Simultaneously, the decoder is connected with the encoder by 1x1 convolution layers. Then, the confidence enhancement module (CEM) takes the prediction of decoder $\mathcal{F}_d $ as input to refine and provide the final co-saliency maps ${\mathcal{M}_1, \mathcal{M}_2}$. At last, the network's output is multiplied with original images $G$ to eliminate the irrelevant regions. Our group-based symmetric triplet (GST) loss is applied on the masked images $G_\mathcal{M}$ to supervise \ourmodel{}~in a metric learning way. Besides, the masked images are then fed to the encoder again to obtain the masked encoded feature $\mathcal{F}_r$. Different from $\mathcal{F}$, $\mathcal{F}_r$ contains only the features of predicted regions and has a more precise semantic representation to be applied in the recurrent auxiliary classification module (RACM) to obtain the classification loss.

\subsection{Group Affinity Module (GAM)}
\label{sec:GAM}

In most cases of real life, objects of the same class share similarity in their appearance and features, which have been widely used in many computer vision tasks. For example, self-supervised video tracking methods~\cite{vondrick2018tracking,lai2019self,wang2019learning,lai2020mast} often propagate the segmentation maps of target objects based on the pixel-wise correspondences between two adjacent frames. Therefore, we introduce this motivation to the CoSOD task by computing the global affinity among all images in the same group.

For features $\{F_{1,n}, F_{1,m}\} \in \mathcal{F}_1$ of any two images\footnote{All  analyses in \secref{sec:GAM} on $\mathcal{F}_1$ can be applied to $\mathcal{F}_2$. 
We omit the group subscript for notation simplicity, \ie{}, we use $F_n$ to represent $F_{1,n}$.}, 
we compute their pixel-wise correlations in the format of an inner product:
\begin{equation}
	S_{(n,m)} =  \theta (F_n)^T \phi(F_m)\label{equation:1},
\end{equation}
where $\theta, \phi$ denote linear embedding functions ($3 \times 3 \times 512$ convolutional layer). The affinity map $S_{(n,m)} \in \mathbb{R}^{HW \times HW}$ efficiently captures the common features of co-salient objects in the given image pair $(n,m)$. Then we can generate $F_n$'s affinity map $A_{n \leftarrow m} \in \mathbb{R}^{HW \times 1}$ by finding the maxima for each of $F_n$'s pixels conditioned on $F_m$, which alleviates the influence of noisy correlation values in the map.

\begin{figure*}[t!]
\centering
\includegraphics[width=.85\linewidth]{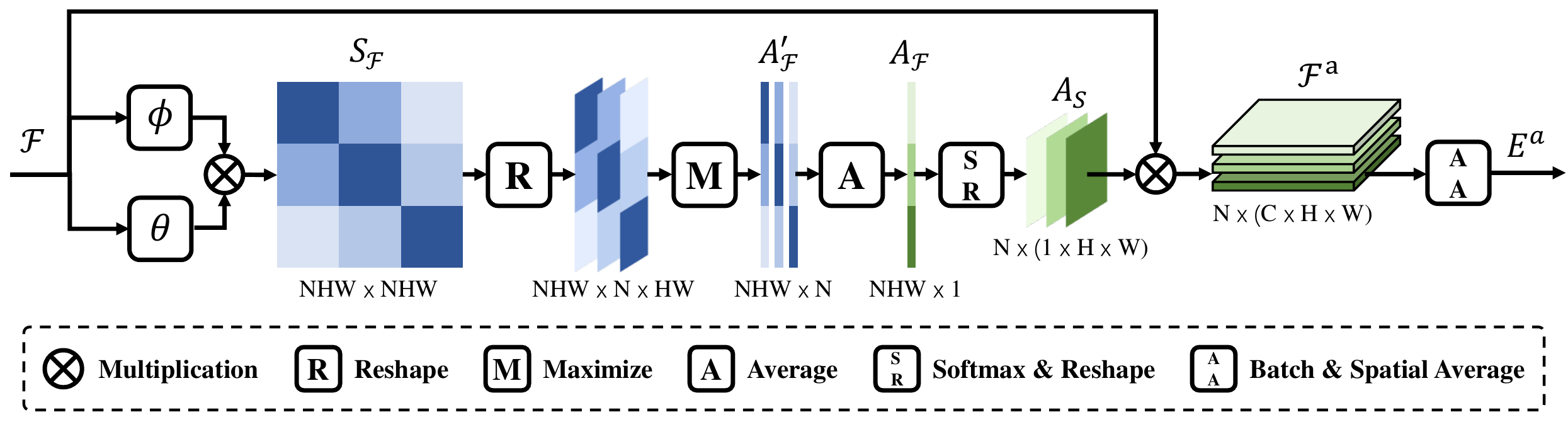}
\caption{\textbf{Group Affinity Module.} We first utilize affinity attention to obtain the attention maps of the input features by collaborating all images in a group. Subsequently, the maps are multiplied with the input features to generate the consensus for the group. Then the obtained consensus is used to coordinate the original feature maps and is also fed to the GCM for inter-group collaborative learning.}
\label{fig:GAM}
\end{figure*}

Similarly, we can extend the use of the local affinity of an image pair to the global affinity of an entire image group. 
Specifically, we compute the affinity map $S_{\mathcal{F}} \in \mathbb{R}^{NHW \times NHW}$ of all image features $\mathcal{F}$ using Eq.~\ref{equation:1}. Then, we find the maxima for each image $A_{\mathcal{F}}' \in \mathbb{R}^{NHW \times N}$ from $S_{\mathcal{F}}$, and average all the maxima of $N$ images to generate the global affinity attention map $A_{\mathcal{F}} \in \mathbb{R}^{NHW \times 1}$. 
In this way, the affinity attention map is globally optimized on all images, and the influence of occasional co-occurring bias is thus alleviated. Then, we use a softmax operation to normalize $A_{\mathcal{F}}$ and reshape it to produce the attention map $A_S \in \mathbb{R}^{N \times (1 \times H \times W)}$. With the attention map $A_S$, we multiply it with the original feature $\mathcal{F}$ to generate the attention feature maps $\mathcal{F}^a \in \mathbb{R}^{N \times C \times H \times W}$.
Finally, the attention feature maps $\mathcal{F}^a$ for the whole group are used to generate the attention consensus $\textit{E}^a$ by the average pooling along both the batch dimension and spatial dimension, as illustrated in \figref{fig:GAM}.

The GAM focuses on capturing the commonality of co-occurring salient objects within the same group, thus improving the intra-group compactness of the consensus representation. Such \textit{intra-group compactness} alleviates the distraction made by co-occurring noise and encourages the model to concentrate on the co-salient regions. 
This allows the shared attributes of co-salient objects to be better captured, resulting in better consensus representation.
The obtained attention consensus $\textit{E}^a$ is combined with the original feature maps $\mathcal{F}$ through depth-wise correlation~\cite{li2019siamrpn++,fan2020few} to achieve efficient information association. The generated feature maps $\mathcal{F}^{out}$ of different groups are then concatenated together and fed to the decoder. After the confidence enhancement module (CEM), final co-saliency maps $\mathcal{M}$ are produced for all images.

\begin{figure}[!t]
\centering
\includegraphics[width=1.0\linewidth]{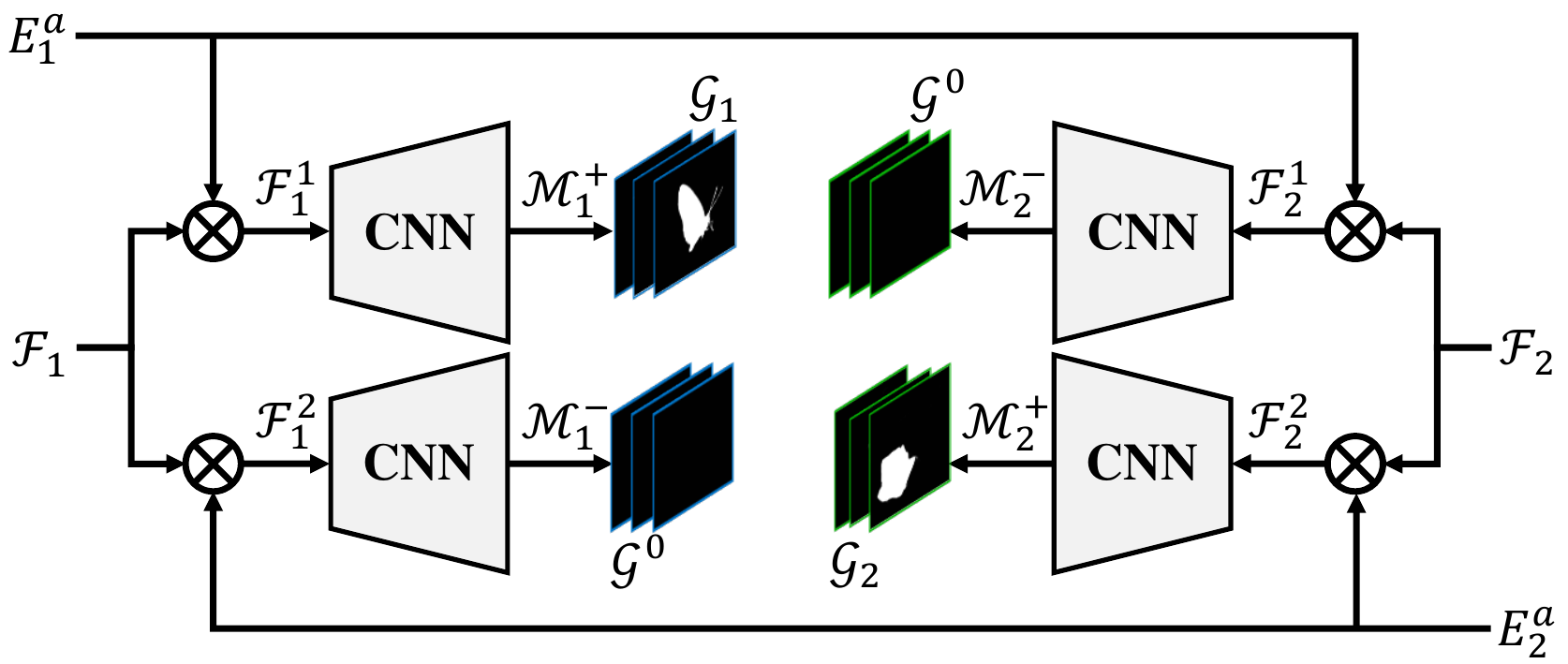}
\caption{\textbf{Group Collaborating Module.} Both groups' original feature maps and consensus are fed to the GCM. The predicted output conditioned on the consistent feature and consensus (from the same group) is supervised with the available ground truth labels. Otherwise, it is supervised by the all-zero maps.}
\label{fig:gcm}
\end{figure}

\subsection{Group Collaborating Module (GCM)}
Currently, most existing CoSOD approaches tend to focus on the intra-group compactness of the consensus. Still, the \textit{inter-group separability} is crucial for distinguishing distracting objects, especially when processing complex images with more than one salient object. 
To this end, we propose a simple but effective module, \ie{}, the GCM, by learning to encode the inter-group separability.

With the GAM, we can obtain the attention consensus $\{\textit{E}_1^a, \textit{E}_2^a\}$ of images in two groups. Then, we apply an intra- and inter-group cross-multiplication ($\cdot$) between the corresponding features $\{\mathcal{F}_1, \mathcal{F}_2\}$ and the attention consensus to get the intra-group collaboration: $\mathcal{F}_1^1 = \mathcal{F}_1 \cdot \textit{E}_1^a$ and $\mathcal{F}_2^2= \mathcal{F}_2 \cdot \textit{E}_2^a$. In contrast, the inter-group multiplication deals with the features and consensus of different groups, \ie{}, $\mathcal{F}_1^2= \mathcal{F}_1 \cdot \textit{E}_2^a$ and $\mathcal{F}_2^1= \mathcal{F}_2 \cdot \textit{E}_1^a$, to represent the inter-group interaction.
The intra-group representation $\mathcal{F}^+ = \{\mathcal{F}_1^1, \mathcal{F}_2^2\}$ is computed to predict co-saliency maps, and the inter-group representation $\mathcal{F}^- = \{\mathcal{F}_1^2, \mathcal{F}_2^1\}$ is employed for a consensus with group separability.
Specifically, we feed the inter-group and intra-group features $\{\mathcal{F}^+, \mathcal{F}^-\}$ to a small convolutional network with an upsampling layer and obtain the saliency maps $\{\mathcal{M}^+, \mathcal{M}^-\}$\footnote{$\mathcal{M}^+ = \{\mathcal{M}_1^+, \mathcal{M}_2^+\}$ and $\mathcal{M}^- = \{\mathcal{M}_1^-, \mathcal{M}_2^-\}$.}
with different supervision signals. As shown in \figref{fig:gcm}, we use the ground truth maps to make supervision on $\mathcal{F}^+$, while all-zero maps on $\mathcal{F}^-$. The loss function is: 
\begin{equation}\label{eqn:GCM}
	{L}_{\text{GCM}} = \frac{1}{N} \sum_n^N {L}_{\text{FL}}(<\mathcal{M}_n^+, \mathcal{M}_n^->, <\mathcal{G}_n, \mathcal{G}_n^0>),
\end{equation}
where ${L}_{\text{FL}}$ denotes the focal loss~\cite{FPN}, $\mathcal{G}_n$ denotes the ground truth, $\mathcal{G}_n^0$ denotes the all-zero map, and $<\cdot>$ denotes the concatenation operation.

Consequently, GCM lets the consensus have a high inter-group separability between different groups and makes it easier to identify distractors in a complex environment. Specifically, this module doesn't introduce additional computation during inference and can be fully discarded.

\begin{figure}[t!]
  \centering
    \begin{overpic}[width=.98\columnwidth]{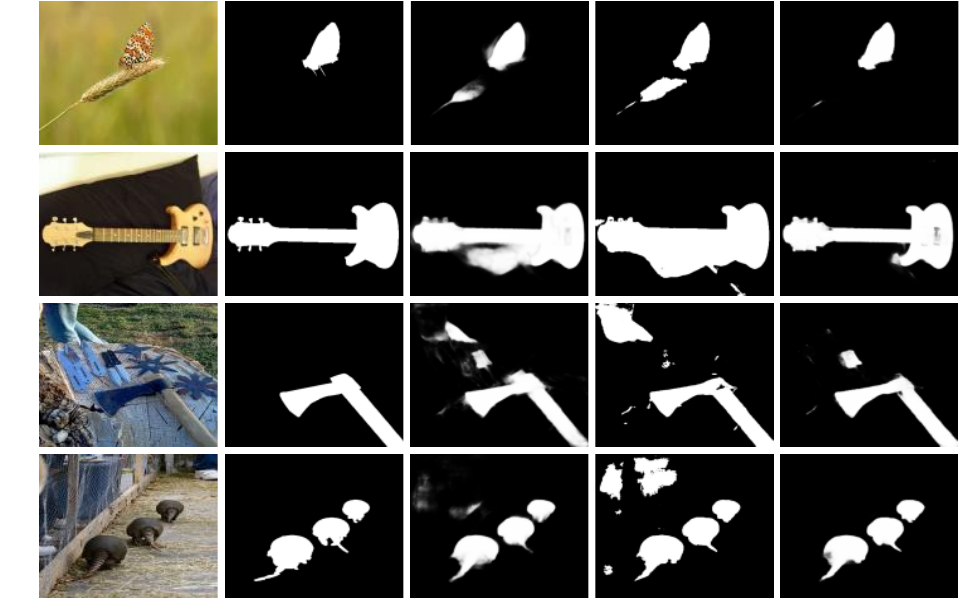}
        \put(0,47.5){\rotatebox{90}{Butterfly}}
        \put(0,33){\rotatebox{90}{Guitar}}
        \put(0,20){\rotatebox{90}{Axe}}
        \put(0,-1){\rotatebox{90}{Armadillo}}
        \put(11.5,-3){{(a)}}
        \put(31,-3){{(b)}}
        \put(50,-3){{(c)}}
        \put(69,-3){{(d)}}
        \put(88.5,-3){{(e)}}
    \end{overpic}
   \caption{\textbf{Prediction results produced by our \ourmodel{}~trained 
   with different losses.} (a) Source image. (b) Ground truth. (c) Results of 
   \ourmodel{}~trained with only BCE loss. (d) Results of \ourmodel{}~trained with only IoU loss. (e) Results of \ourmodel{}~trained with balanced BCE and IoU loss. All the results here are generated from models trained with the DUTS\_class dataset.}
   \label{fig:BCE_vs_IoU}
\end{figure}
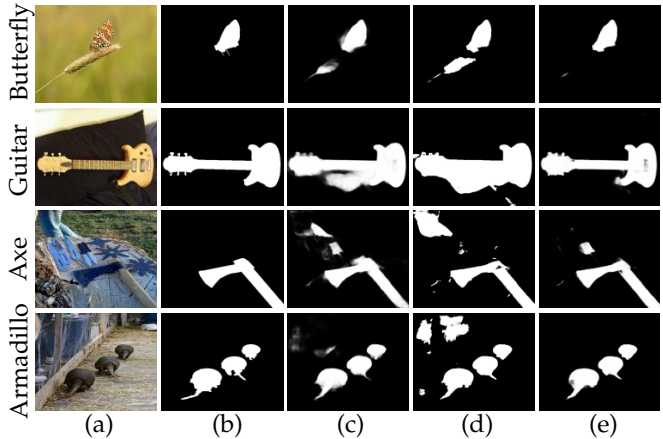

\subsection{Confidence Enhancement Module (CEM)}
In SOD tasks, the pixel values of predicted saliency maps range between 0 and 1 since the network is usually appended with a Sigmoid function. Although the pixel values of ground truth maps are either 0 or 1, those of the predicted saliency maps may approach 0.5, which indicates more uncertainty and noise in the prediction. On the very hard cases, results with more uncertainty and noise tend to get higher scores by some classic metrics~\cite{Emeasure},~\eg{}, Fbw~\cite{Fbw}, IoU~\cite{IoU}, MAE,~\etc{}, while terrible in practice, which is opposite to the final goal.

To deal with the uncertain values of the prediction, we conduct research from both the perspective of the loss function and network architecture.
From the view of the loss function, we set up comparative experiments to verify that different loss functions can introduce different optimization directions to the same network. To be more specific, IoU loss guides the outputs to be almost 0 or 1, but with the low accuracy of existing metrics, \ie{}, S-measure~\cite{Smeasure}, E-measure~\cite{Emeasure}. In contrast, BCE loss directs the network to predict more uncertain values but achieve better scores in the above metrics. As the expected maps are shown in \figref{fig:BCE_vs_IoU}, although IoU loss brings high confidence to predicted maps, the optimization is too rough. It acts up in terms of the integrity of the saliency maps. Therefore, BCE loss is still a necessity for training. To improve the quality of saliency maps in terms of binarization for practical application, we try to balance the BCE and IoU loss as a mixed pixel loss for supervision.

\begin{figure}[t!]
    \centering
    \begin{overpic}[width=.99\columnwidth]{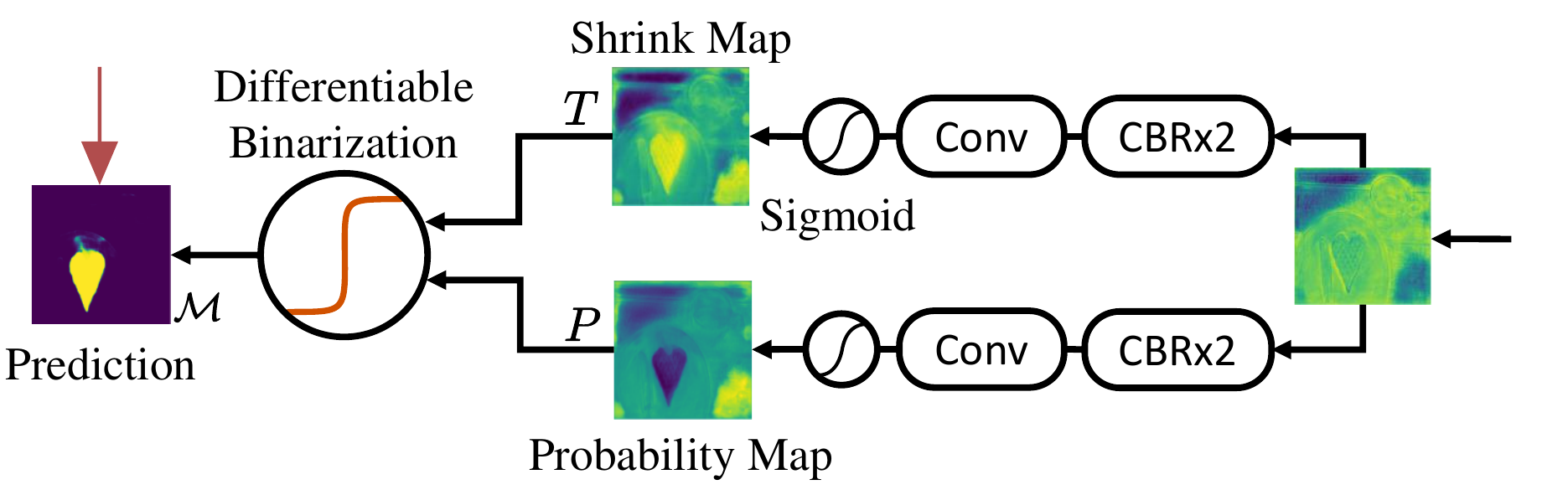}
        \put(1,29){\footnotesize {${L}_{\text{BCE}} + {L}_{\text{IoU}}$}}
        \put(96,15){{$\mathcal{F}_d$}}
    \end{overpic}
    \vspace{-10pt}
    \caption{\textbf{Confidence Enhancement Module.} After the decoder, we adapt the CEM to bring higher quality and binarization to the predicted saliency maps. CBR means a convolution layer followed by a batch normalization layer and a ReLU activation function.}
    \label{fig:CEM}
\end{figure}

From the view of network architecture, we employ the confidence enhancement module (CEM) at the end of \figref{fig:Arch}. In previous SOD approaches, the Sigmoid function is usually applied to squeeze the output values from 0 to 1. However, as is described in~\cite{DB}, the Sigmoid activation function is not steep enough, and the values produced by it are not binarized enough.
To address this issue, as shown in \figref{fig:CEM}, the output feature $\mathcal{F}_d$ of the decoder is fed into the CEM. Firstly, the feature $\mathcal{F}_d$ goes through two parallel branches with two 3x3 convolution layers, which are both followed by batch normalization, a ReLU activation function, and a 1x1 convolution layer followed by a Sigmoid activation function. After that, the probability map $P$ and the threshold map $T$ are generated and put into the differentiable binarization function to obtain the final prediction.
According to~\cite{DB}, the final co-saliency maps $\mathcal{M}$ can be represented as:
\begin{equation}
  \mathcal{M}_{i, j} = \frac{1}{1 + e ^ {-k (P_{i, j} - T_{i, j})}},
  \label{eqn:DB} 
\end{equation}
where $k$ is the factor that controls the steepness of the step function. In our implementation, $k$ is set to 300 as the default value. When loss meets NaN during training, it will be replaced with 50 for current propagation.

\subsection{Group-based Symmetric Triplet (GST) Loss}
\label{sec:GST}
In the past few years, some approaches have been designed to solve the CoSOD task from the perspective of metric learning~\cite{MetricCoSOD, MetricCoSOD1}. However, most existing CoSOD approaches based on metric learning use super-pixel~\cite{SuperPixel} to extract fractions as the unit of measurement. Most of these methods are usually not end-to-end and in low efficiency.
Besides, existing works typically introduce class labels to help the model learn more representative features with high semantics. Specifically, in~\cite{GICD}, Zhang~\etal{}divide the DUTS dataset~\cite{DUTS} into different groups by the class of main salient objects to build the training set. However, absolute class labels may not be given in realistic scenarios. In contrast, there is only the relative label of two groups (whether they belong to the same group).
In 2015, Schroff~\etal{}proposed the triplet loss~\cite{TripletLoss} to help face recognition, which is a good way to learn the discriminative feature of different identities by pulling the positive samples and pushing negative samples.
Because of the success of triplet loss in face recognition~\cite{TripletLoss}, visual tracking~\cite{TripletLoss_OT}, person re-id~\cite{TripletLoss_PR},~\etc{}, we modify the original triplet loss to GST loss to learn more discriminative features from different groups, which could improve the uniqueness and discrimination of the consensus features of objects with different class labels.

\begin{figure}[t!]
  \centering
   \begin{overpic}[width=0.98\linewidth]{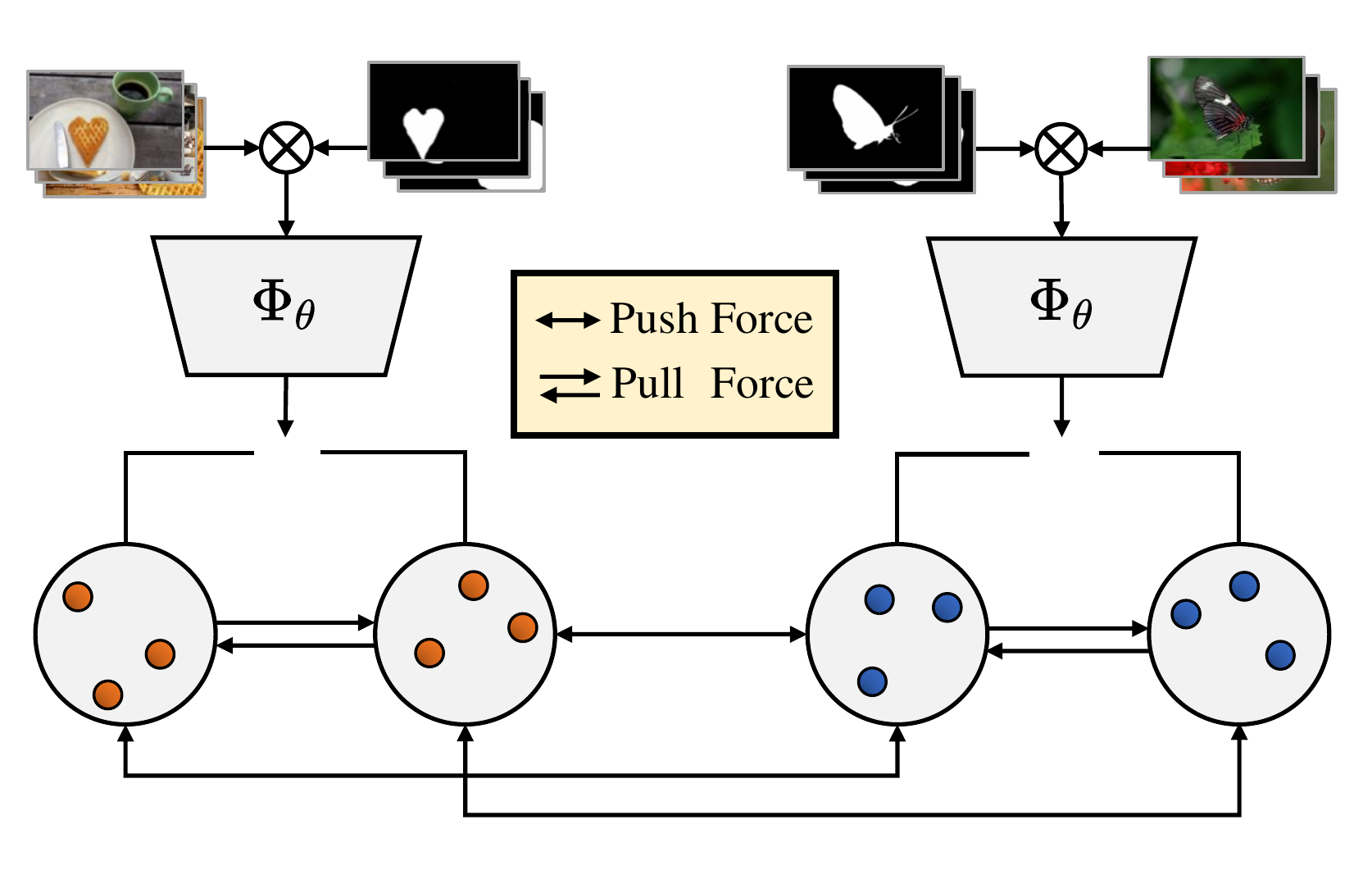}
       \put(6,47){\footnotesize {$G_1$}}
       \put(32,47){\footnotesize {$\mathcal{M}_1$}}
       \put(62,47){\footnotesize {$\mathcal{M}_2$}}
       \put(88,47){\footnotesize {$G_2$}}
       \put(1,26){\footnotesize {$\mathcal{F}_r^{1A}$}}
       \put(25,25){\footnotesize {$\mathcal{F}_r^{1B}$}}
       \put(66,26){\footnotesize {$\mathcal{F}_r^{2A}$}}
       \put(83,25){\footnotesize {$\mathcal{F}_r^{2B}$}}
       \put(18.5,30.5){\footnotesize {$\mathcal{F}_r^1$}}
       \put(75,30.5){\footnotesize {$\mathcal{F}_r^2$}}
       \put(48,2){\footnotesize {Euclidean~Distance}}
       
   \end{overpic}
   \vspace{-10pt}
   \caption{\textbf{Group-based Symmetric Triplet Loss.} In GST loss, each group is divided averagely into two sub-groups. The sub-groups from the same group pull to each other and push to features of other groups. $\Phi_{\theta}$ represents the backbone (see \figref{fig:Arch}).}
   \label{fig:loss_GST}
\end{figure}
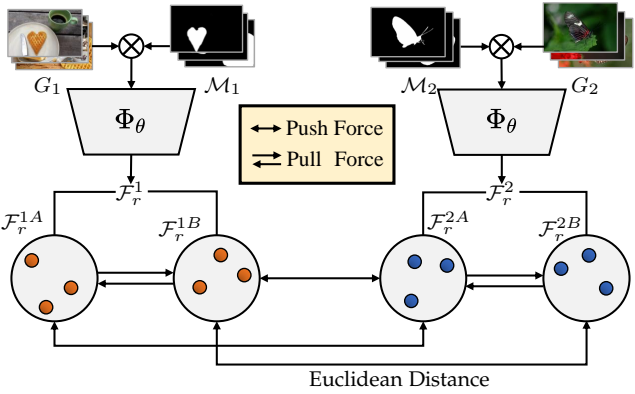

Note that our GST loss is only activated in the training process. Specifically, it is applied on $\mathcal{F}_r$, which is the output feature extracted by the encoder from $G_\mathcal{M}$, \ie{}, the multiplication result between predicted saliency maps $\mathcal{M}$ and the original images $G$ (see \figref{fig:Arch}). In this way, only the pixels of targeted objects are used for the measurement. Taking $G_1$ in \figref{fig:loss_GST} as an example, the backbone $\Phi_{\theta}$ extracts the semantic representation $\mathcal{F}_r^1$ from the raw images masked with $\mathcal{M}_1$. Then, the $\mathcal{F}_r^1$ is split into two parts by class, \ie{}, $\mathcal{F}_r^{1A}$, $\mathcal{F}_r^{2A}$. Features from the same group are seen as positive samples of each other, while those from the other group are negative. As shown in \figref{fig:loss_GST}, our GST loss is calculated in a symmetric structure. Finally, the triplet loss is computed on both the $(\mathcal{F}_r^{1A},~\mathcal{F}_r^{1B},~\mathcal{F}_r^{2A})$ and $(\mathcal{F}_r^{1B},~\mathcal{F}_r^{2B},~\mathcal{F}_r^{2A})$, where the distances between features are measured with Euclidean distance. Specifically, ${L}_\text{Tri}(\mathcal{F}_r^{1A},~\mathcal{F}_r^{1B},~\mathcal{F}_r^{2A})$ can be denoted as follows:
\begin{equation}
  ||\mathcal{F}_r^{1A} - \mathcal{F}_r^{1B}||_2 - ||\mathcal{F}_r^{1B} - \mathcal{F}_r^{2A}||_2 + \alpha,
  \label{eqn:loss_tri}
\end{equation}
where $\alpha$ denotes the margin that is a hyper-parameter enforced between positive and negative pairs~\cite{TripletLoss}. $||\cdot||_2$ denotes the two-norm of input. Because of the symmetry of GST loss, ${L}_\text{Tri}(\mathcal{F}_r^{1B},~\mathcal{F}_r^{2B},~\mathcal{F}_r^{2A})$ is also measured with the Euclidean distance in the same way.

The final GST loss is a combination of double $L_\text{Tri}$ when $G_1$ and $G_2$ act as the positive samples alternately from the images masked with predicted maps:
\begin{equation}
  {L}_\text{GST} = {L}_\text{Tri}(\mathcal{F}_r^{1A},~\mathcal{F}_r^{1B},~\mathcal{F}_r^{2A})~+~{L}_\text{Tri}(\mathcal{F}_r^{1B},~\mathcal{F}_r^{2B},~\mathcal{F}_r^{2A}),
  \label{eqn:loss_GST}
\end{equation}

\subsection{Recurrent Auxiliary Classification Module (RACM)}\label{sec:RACM}
Existing works typically train the model with images within the same group to extract common information. Specifically, images in a certain batch only have ground truth maps on the objects belonging to the same class, where only the common intra-group features can be learned.
However, since there are no constraints on the features learned, common features of different classes may get close to each other and be hard to distinguish.

In~\cite{GCoNet}, the auxiliary classification module (ACM) facilitates the high-level semantic representation to obtain more discriminative features for consensus learning. Specifically, a class predictor consisting of a global average pooling layer and one fully connected layer is applied after the backbone. The features of objects with the same class are clustered together through class-level supervision.
Although ACM works well in GCoNet~\cite{GCoNet}, it has some defects: the features from the backbone are unstable and can be something else other than the right objects. As a consequence, the ACM may give a wrong optimizing direction. Meanwhile, it runs implicitly and is difficult to monitor.

We propose to use the Recurrent ACM (RACM) to overcome the problems mentioned above. The pipeline of RACM is kept almost the same as that of the original ACM. In contrast, RACM takes the model's output as the mask to obtain the pixels of target objects only rather than the whole image used in ACM. Then the masked images will be sent again to the encoder and class predictor. After eliminating other distracting regions, our RACM focuses only on the interesting areas. When the prediction of our \ourmodel{}~is far from the ground truth map, RACM can give an enhanced penalty to help accelerate the convergence of training. Combining the raw images and ground truth maps to formulate the loss, RACM enables the model to learn more discriminative features for inter-group separability and intra-group compactness, respectively. 
The loss functions of classification are as follows:
\begin{equation}
  \hat{Y}_\text{ACM} = \varphi(\Phi_{\theta}(G)),
  \label{eqn:loss_acm}
\end{equation}
\begin{equation}
  \hat{Y}_\text{RACM} = \varphi(\Phi_{\theta}(G \otimes \mathcal{M})),
  \label{eqn:loss_racm}
\end{equation}
\begin{equation}
  {L}_\text{CLS} = {L}_\text{CE}(\hat{Y}_\text{RACM},~Y_\text{CLS}) + {L}_\text{CE}(\hat{Y}_\text{ACM},~Y_\text{CLS}),
  \label{eqn:loss_cls}
\end{equation}
where $\varphi$ and $\Phi_{\theta}$ denote the class predictor (GAP and one linear layer) and encoder, respectively. $L_\text{CE}$ is the cross-entropy loss, $Y_\text{CLS}$ are the ground truth class labels, and $\hat{Y}_\text{ACM}$ and $\hat{Y}_\text{RACM}$ are the class labels predicted by ACM and RACM, respectively. 

\subsection{Objective Function}
The objective function is a weighted combination of saliency map loss (combination of BCE loss and IoU loss), GCM loss, our GST loss, and classification loss.
The BCE loss and IoU loss are illustrated as:
\begin{equation}
  {L}_\text{BCE} = - \sum{[Y log(\hat{Y}), (1 - Y)log(1 - \hat{Y})]},
  \label{eqn:loss_BCE}
\end{equation}
\begin{equation}
  {L}_\text{IoU} = 1 - \frac{1}{N}\sum{\frac{Y \cap \hat{Y}}{Y \cup \hat{Y}}},
  \label{eqn:loss_IoU}
\end{equation}
where $Y$ is the ground truth and $\hat{Y}$ is the prediction.
With the GCM loss~\eqnref{eqn:GCM}, GST loss~\eqnref{eqn:loss_GST}, and classification loss~\eqnref{eqn:loss_cls}, 
our final objective function is:
\begin{equation}
  {L} = \lambda_1 {L}_\text{BCE} + \lambda_2 {L}_\text{IoU} + \lambda_3 {L}_\text{GCM} + \lambda_4 {L}_\text{GST} + \lambda_5 {L}_\text{CLS},
  \label{eqn:loss_final}
\end{equation}
where $\lambda_1,~\lambda_2,~\lambda_3,~\lambda_4,~and~\lambda_5$ are respectively set to 30, 0.5, 250, 3, and 3 to keep all the losses on the same quantitative level at the beginning of training.


\section{Experiments}
This section provides the guidelines and details in our base and extensive experiments, \ie{}, datasets, settings, evaluation protocol, and analysis in training and testing, respectively.

\subsection{Datasets}
\label{sec:datasets}
\textbf{Training Sets.}
We follow the GICD~\cite{GICD} to use DUTS\_class as our training set to design the experiments. After removing the noisy samples by Zhang~\etal{}~\cite{GICD}, the whole DUTS\_class is divided into 291 groups, which contain 8,250 images in total. The DUTS\_class dataset is the only training set used for evaluation in our ablation study. 
Nowadays, there is still a lack of a fully recognized training dataset. To make a fair comparison with  up-to-date works~\cite{GWD,COCO_SEG,ICNet,CADC,CoADNet}, we  employed the widely-adopted COCO-9k~\cite{GWD}, a subset of the COCO~\cite{COCO} with 9,213 images of 65 groups, and the COCO-SEG~\cite{COCO_SEG} which is also a subset of the COCO~\cite{COCO} and contains 200k images, to train our \ourmodel{}~as supplementary experiments.

\textbf{Test Sets.}
To obtain a comprehensive evaluation of our \ourmodel{}, we test it on three widely used CoSOD datasets, \ie{}, CoCA~\cite{GICD}, CoSOD3k~\cite{CoSOD3k}, and CoSal2015~\cite{CoSal2015}. Among these three datasets, CoCA is the most challenging dataset. 
It is of much higher diversity and complexity in terms of background, occlusion, illumination, surrounding objects,~\etc{}. Following the latest benchmark~\cite{CoSOD3k}, we do not evaluate on iCoseg~\cite{iCoseg} and MSRC~\cite{MSRC}, since only one salient object is given in most images there. It is more convincing to evaluate CoSOD methods on images with more salient objects, which is closer to real-life applications.

\subsection{Evaluation Protocol}
Following the GCoNet~\cite{GCoNet}, we employ the S-measure~\cite{Smeasure}, maximum F-measure~\cite{Fmeasure}, maximum E-measure~\cite{Emeasure}, and mean absolute error (MAE) to evaluate the performance in our experiments. Evaluation toolbox can be referred to \url{https://github.com/zzhanghub/eval-co-sod}.

\textbf{S-measure}~\cite{Smeasure} is a structural similarity measurement between a saliency map and its corresponding ground truth map. The evaluation with $S_{\alpha}$ can be obtained at high speed without binarization. S-measure is computed as:
\begin{equation}
  S_{\alpha} = \alpha \times S_o + (1 - \alpha) \times S_{r},
  \label{eqn:Sm}
\end{equation}
where $S_o$ and $S_r$ denote object-aware and region-aware structural similarity, and $\alpha$ is set to 0.5 by default, as suggested by Fan~\etal{}in~\cite{GCoNet}.

\textbf{F-measure}~\cite{Fmeasure} is designed to evaluate the weighted harmonic mean value of precision and recall. The output of the saliency map is binarized with different thresholds to obtain a set of binary saliency predictions. The predicted saliency maps and ground truth maps are compared for precision and recall values. The best F-measure score obtained with the best threshold for the whole dataset is defined as $F_{\beta}^{max}$. F-measure can be computed as:
\begin{equation}
  F_{\beta} = \frac{(1+\beta^2) Precision \times Recall}{\beta^2 (Precision + Recall)},
  \label{eqn:Fm}
\end{equation}
where $\beta^2$ is set to 0.3 to emphasize the precision over recall, following~\cite{SOD_review1}.

\textbf{E-measure}~\cite{Emeasure} is designed as a perceptual metric to evaluate the similarity between the predicted maps and ground truth maps from both local and global views. 
E-measure is defined as:
\begin{equation}
  E_{\xi} = \frac{1}{W \times H}\sum_{x=1}^{W}\sum_{y=1}^{H}\phi_{\xi}(x, y),
  \label{eqn:Em}
\end{equation}
where $\phi_{\xi}$ indicates the enhanced alignment matrix. Similar to the F-measure, we also adopt the max E-measure ($E_{\xi}^{max}$) as our evaluation metrics.

\begin{table*}
\begin{center}
\footnotesize
\renewcommand{\arraystretch}{1.0}
\renewcommand{\tabcolsep}{2mm}
\caption{\textbf{Quantitative ablation studies of the overall modification on the framework of our \ourmodel{}.} We conduct the ablation studies of our \ourmodel{}~on the effectiveness of overall modification on the framework, including network simplification (Net-Sim), batch normalization (BN), and hybrid loss (HL). }
\label{table:ablation_modif_overall}
\begin{tabular}{c|ccc||cccc|cccc|cccc}
\hline
& \multicolumn{3}{c||}{Modules}  & \multicolumn{4}{c|}{CoCA~\cite{GICD}} & \multicolumn{4}{c|}{CoSOD3k~\cite{CoSOD3k}} & \multicolumn{4}{c}{CoSal2015~\cite{CoSal2015}} \\
ID & \hspace{0.25mm}Net-Sim\hspace{0.25mm} & BN & HL & $E_{\xi}^\text{ max} \uparrow$ & $S_\alpha \uparrow$ & $F_\beta^\text{ max} \uparrow$ & $\epsilon \downarrow$ & $E_{\xi}^\text{ max} \uparrow$ & $S_\alpha \uparrow$ & $F_\beta^\text{ max} \uparrow$ & $\epsilon \downarrow$ & $E_{\xi}^\text{ max} \uparrow$ & $S_\alpha \uparrow$ & $F_\beta^\text{ max} \uparrow$ & $\epsilon \downarrow$ \\
\hline
1 &  &  &  & 0.760 & 0.673 & 0.544  & 0.105 & 0.860 & 0.802 & 0.777  & 0.071 & 0.888 & 0.845 & 0.847  & 0.068 \\
2 & \checkmark &  &  & 0.752 & 0.676 & 0.538 & \textbf{0.100} & 0.872 & 0.815 & 0.796 & 0.063 & 0.895 & 0.853 & 0.858 & 0.063 \\
3 & \checkmark & \checkmark &  & 0.747 & 0.683 & 0.556 & 0.110 & \textbf{0.884} & 0.824 & 0.806 & \textbf{0.062} & 0.912 & 0.868 & 0.874 & \textbf{0.051} \\
4 & \checkmark &  & \checkmark & 0.774 & \textbf{0.691} & \textbf{0.562} & 0.106 & 0.879 & \textbf{0.831} & 0.806 & 0.065 & 0.901 & 0.867 & 0.865 & 0.062 \\
\hline
\rowcolor{mygray}
5 & \checkmark & \checkmark & \checkmark & \textbf{0.779} & 0.681 & 0.558  & 0.119 & 0.882  & 0.828  & \textbf{0.807}  & 0.068 & \textbf{0.913} & \textbf{0.875} & \textbf{0.877} & 0.055 \\
\hline
\end{tabular}
\end{center}
\end{table*}

\textbf{MAE} $\epsilon$ is a simple pixel-level evaluation metric that measures the absolute difference between the predicted maps and the ground truth maps without binarization.
It is defined as:
\begin{equation}
  \epsilon = \frac{1}{W \times H}\sum_{x=1}^{W}\sum_{y=1}^{H}|\hat{Y}(x,y) - \text{GT}(x, y)|.
  \label{eqn:MAE}
\end{equation}

\subsection{Implementation Details}
Based on GCoNet~\cite{GCoNet}, we employ VGG-16 with batch normalization~\cite{BN} as the backbone. We randomly pick $N$ samples from two different groups in each training batch.

\begin{equation}
  N = \min(\#group A, \#group B, 32),
  \label{eqn:batch_size}
\end{equation}
where $N$ denotes the batch size for training, and $\#$ means the number of images in the corresponding group.
Due to the small number of images in some groups, we chose the minimum between 32 and the smaller number of images in the two groups which we randomly selected. Note that $N$ for training and testing can be different. During testing, we follow previous works~\cite{GCoNet,GICD,CADC,CoADNet,DCFM,DeepACG} to set the exact number of images in the given group as the batch size $N$.

To clarify our proposed network, we provide the hyper-parameters in the newly proposed modules. The steep step function produces some NaN values after the backpropagation in the confidence enhancement module (CEM). So, we set the $k$ in differentiable binarization (DB) to a radical value of 300 and a conservative value of 50. When NaN is produced in a certain step, 50 will be used for replacement, which never produces NaN in our experiments. In group-based symmetric triplet (GST) loss, the margin value is set to 1.0.

The images are resized to 256x256 for training and testing. 
The output maps are resized to the original size for evaluation. 
Three data augmentation strategies are applied in our training process, \ie{}, horizontal flip, color enhancement, and rotation. Our \ourmodel{}~is trained over 320 epochs with the Adam optimizer. The initial learning rate is set to 3e-4, $\beta_1 = 0.9$, and $\beta_2 = 0.99$. The whole training process takes around 20 hours. All the experiments are implemented based on PyTorch~\cite{PyTorch} with a single Tesla V100 GPU.

\begin{figure}[t!]
	\centering
    \includegraphics[width=\linewidth]{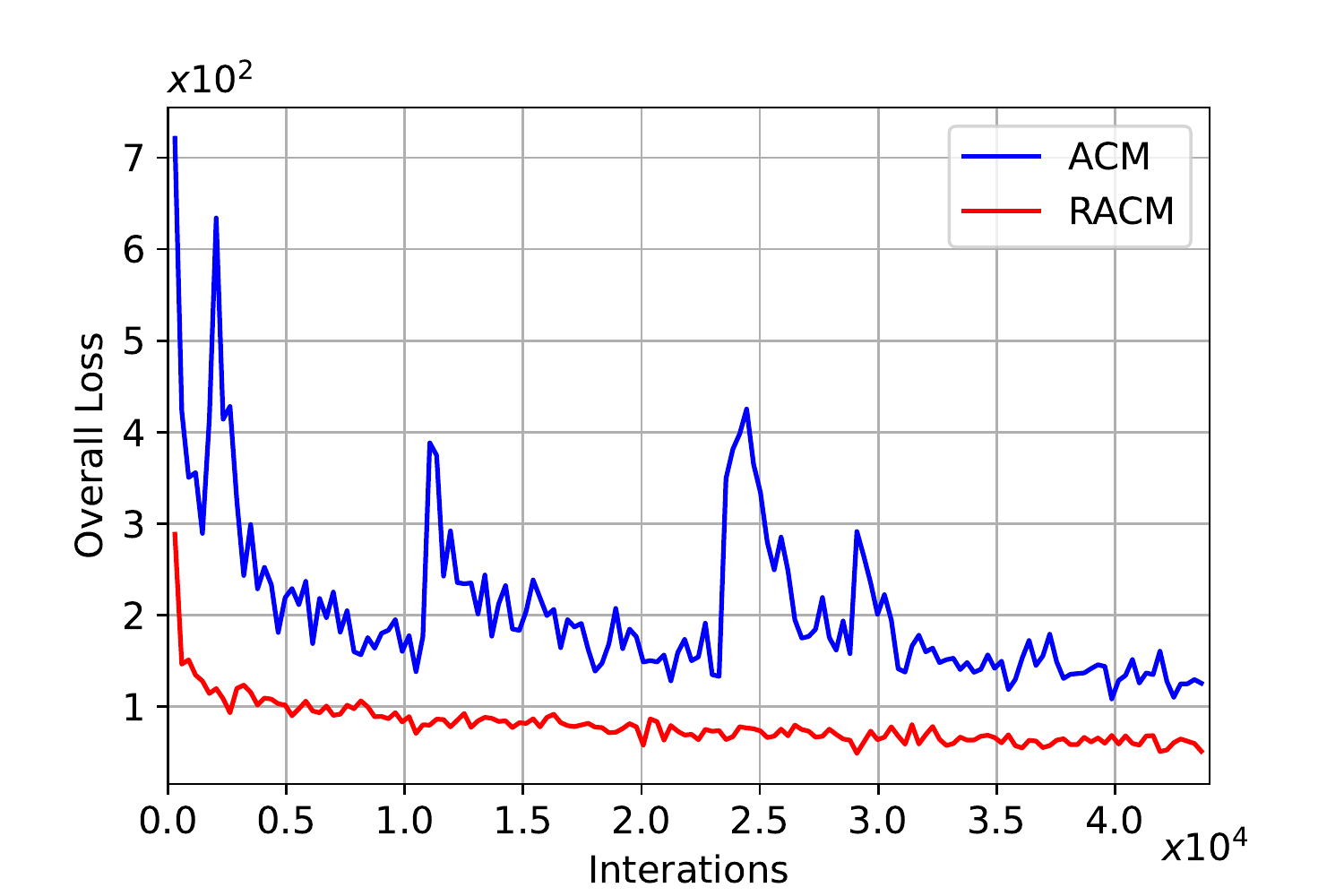}
    \vspace{-20pt}
	\caption{\textbf{Learning curve comparison}. We record the overall losses obtained during the training of our baseline (see \secref{sec:Ablation}) with additional RACM and with only original ACM, where DUTS\_class is used as the training set.}
	\label{fig:racm_acm_lossTR}
\end{figure}

\subsection{Ablation Study}
\label{sec:Ablation}
\begin{figure*}[t!]
	\centering
    \begin{overpic}[width=.97\textwidth]{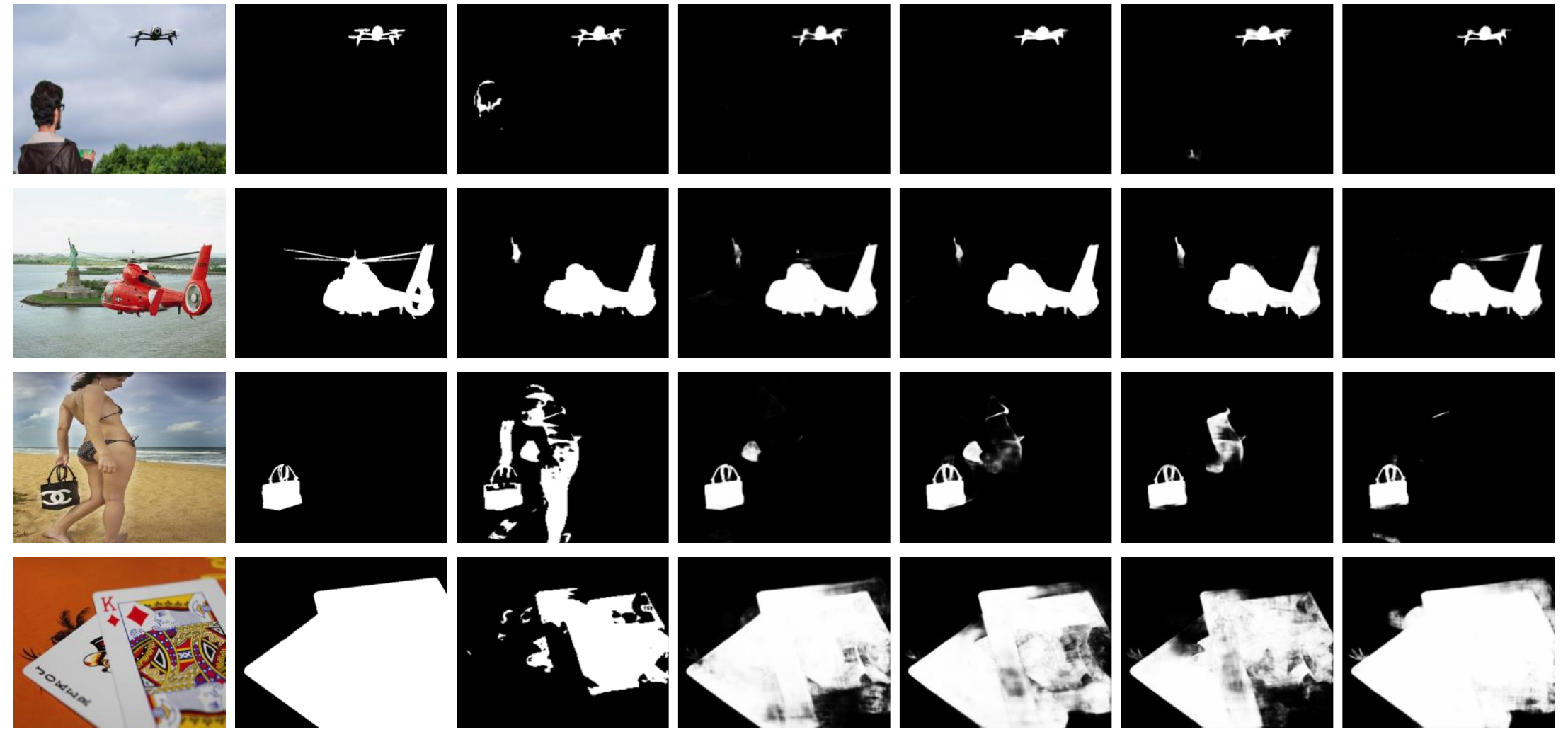}
        \put(-1.5,39){\rotatebox{90}{UAV}}
        \put(-1.5,25){\rotatebox{90}{Helicopter}}
        \put(-1.5,14){\rotatebox{90}{Handbag}}
        \put(-1.5,3.5){\rotatebox{90}{Poker}}
        \put(7,-2){{(a)}}
        \put(21,-2){{(b)}}
        \put(35,-2){{(c)}}
        \put(49,-2){{(d)}}
        \put(63,-2){{(e)}}
        \put(77.5,-2){{(f)}}
        \put(91.5,-2){{(g)}}
    \end{overpic}
	\caption{\textbf{Qualitative ablation studies of our \ourmodel{}~on different modules and their combinations.} (a) Source image; (b) Ground truth; (c) GCoNet~\cite{GCoNet}; (d) Our new baseline; (e) Baseline+RACM; (f) Baseline+RACM+CEM; (g) Baseline+RACM+CEM+GST, the final version of our \ourmodel{}. To keep consistency with GCoNet, the predicted maps provided here are generated by our \ourmodel{}~trained on DUTS\_class only.}
	\label{fig:qualitive_ablation}
\end{figure*}
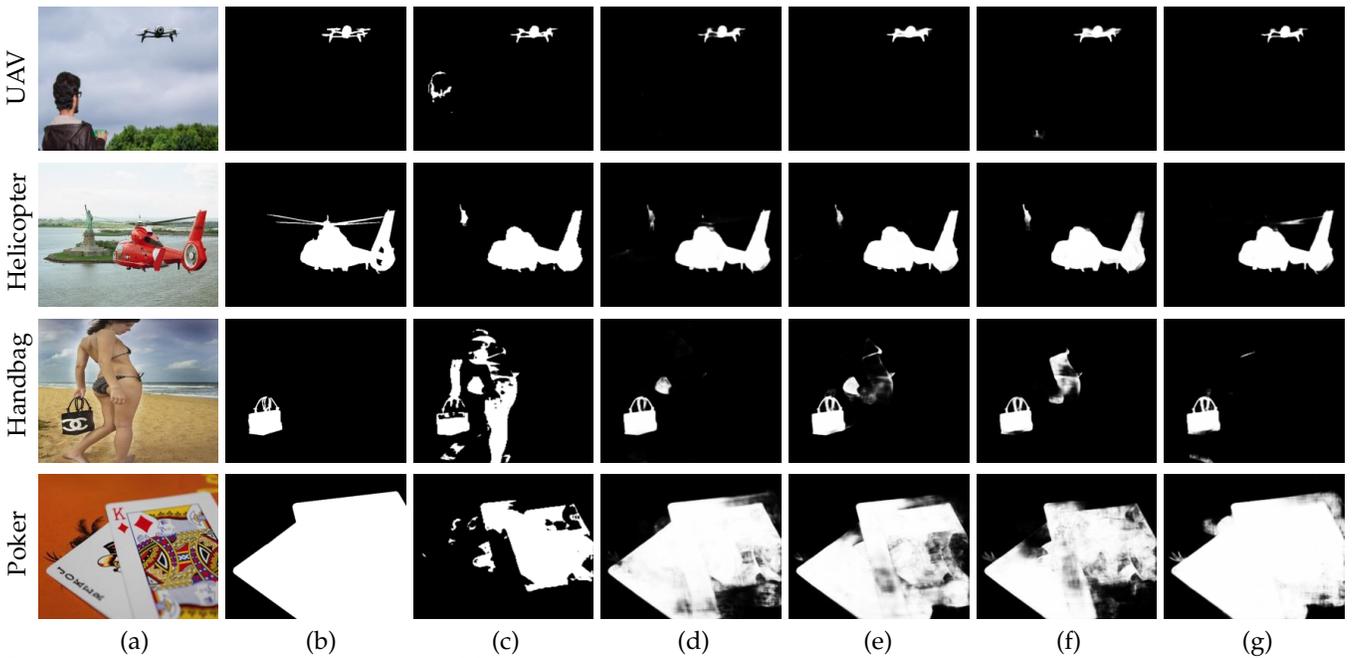

We study the effectiveness of each extension component (\ie{}, RACM, CEM, and GST) employed in our \ourmodel{}~and investigate why they can help learn both good consensus features and discriminative features in our framework.
The qualitative results regarding each module are shown in \figref{fig:qualitive_ablation}. For more ablation studies and experimental settings, can be referred to our conference version~\cite{GCoNet}.

\begin{table*}[t!]
\begin{center}
\footnotesize
\renewcommand{\arraystretch}{1.0}
\setlength\tabcolsep{5.3pt}
\caption{\textbf{Quantitative ablation studies of the proposed components in our \ourmodel{}.} We conduct the ablation studies of our \ourmodel{}~on the effectiveness of the proposed components, including RACM (Recurrent Auxiliary Classification Module), CEM (Confidence Enhancement Module), GST (Group-based Symmetric Triplet Loss), and their combinations.}
\label{table:ablation_modules}
\begin{tabular}{c|ccc||cccc|cccc|cccc}
\hline
& \multicolumn{3}{c||}{Modules}  & \multicolumn{4}{c|}{CoCA~\cite{GICD}} & \multicolumn{4}{c|}{CoSOD3k~\cite{CoSOD3k}} & \multicolumn{4}{c}{CoSal2015~\cite{CoSal2015}} \\
ID &  \hspace{1.25mm}RACM\hspace{1.25mm} & CEM & GST & $E_{\xi}^\text{ max} \uparrow$ & $S_\alpha \uparrow$ & $F_\beta^\text{ max} \uparrow$ & $\epsilon \downarrow$ & $E_{\xi}^\text{ max} \uparrow$ & $S_\alpha \uparrow$ & $F_\beta^\text{ max} \uparrow$ & $\epsilon \downarrow$ & $E_{\xi}^\text{ max} \uparrow$ & $S_\alpha \uparrow$ & $F_\beta^\text{ max} \uparrow$ & $\epsilon \downarrow$ \\
\hline
1 &  &  &  & 0.779 & 0.681 & 0.558 & 0.119 & 0.882 & 0.828 & 0.807 & 0.068 & 0.913 & 0.875 & \textbf{0.877} & 0.055 \\
2 & \checkmark &  &  & 0.780 & 0.684 & 0.570 & 0.120 & \textbf{0.884} & 0.829 & 0.809 & \textbf{0.067} & 0.912 & 0.873 & 0.875 & 0.056 \\
3 & \checkmark & \checkmark &  & 0.779 & 0.686 & 0.565 & 0.117 & 0.881 & 0.829 & 0.805 & 0.068 & 0.913 & 0.872 & 0.873 & 0.057 \\
4 & \checkmark &  & \checkmark & 0.780 & 0.683 & 0.559 & 0.118 & 0.882 & \textbf{0.831} & \textbf{0.810} & 0.068 & 0.914 & \textbf{0.876} & 0.876 & 0.055 \\
\hline
\rowcolor{mygray}
5 & \checkmark & \checkmark & \checkmark & \textbf{0.786} & \textbf{0.691} & \textbf{0.574} & \textbf{0.113} & 0.881 & 0.828 & 0.807 & 0.068 & \textbf{0.917} & 0.875 & 0.876 & \textbf{0.054} \\
\hline
\end{tabular}
\end{center}
\end{table*}

\begin{figure*}[t!]
	\centering
    \begin{overpic}[width=.97\textwidth]{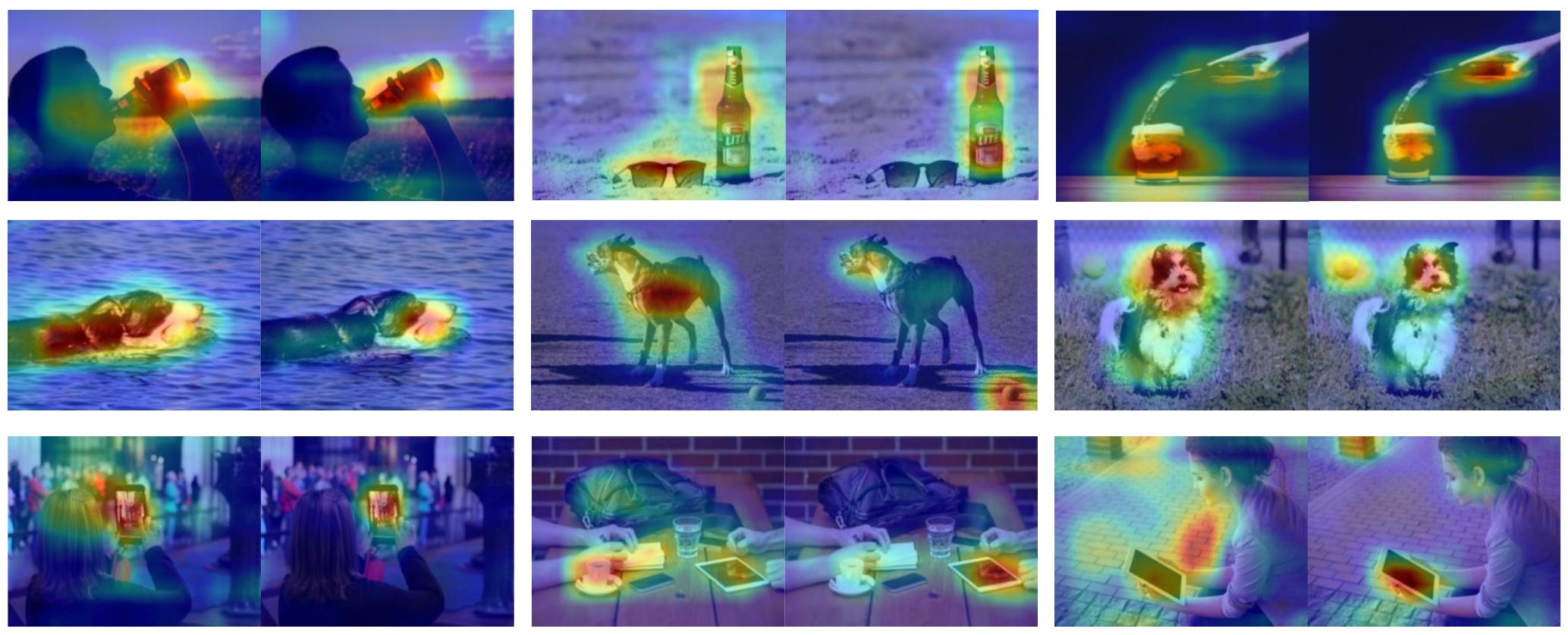}
        \put(-1.5,29.5){\rotatebox{90}{Beer Bottle}}
        \put(-1.5,16){\rotatebox{90}{Tennis Ball}}
        \put(-1.5,4){\rotatebox{90}{Tablet}}
        \put(16,-2){{(a)}}
        \put(49,-2){{(b)}}
        \put(82.5,-2){{(c)}}
    \end{overpic}
	\caption{
    \textbf{The class activation maps~\cite{CAM} are obtained on the classification branch.} For each comparison cell, the left half results are the activation maps obtained from the original GCoNet~\cite{GCoNet} using ACM; the right half results are the activation maps generated by our \ourmodel{}~with extra RACM. As column (a) shows, our \ourmodel{}~has superiority in precisely putting its attention on the target object next to other objects around. In (b), our \ourmodel{}~shows better performance in focusing on objects of the correct class though it has some disturbing surroundings. In the last column (c), some complex samples make both models mistaken. Although some wrong attention is put on objects of the wrong classes, our \ourmodel{}~can still put most of the attention on the right objects and see them as the main parts of the images. The classification activation maps provided here are produced by our \ourmodel{}~trained on DUTS\_class only.
	}\label{fig:CAM}
\end{figure*}
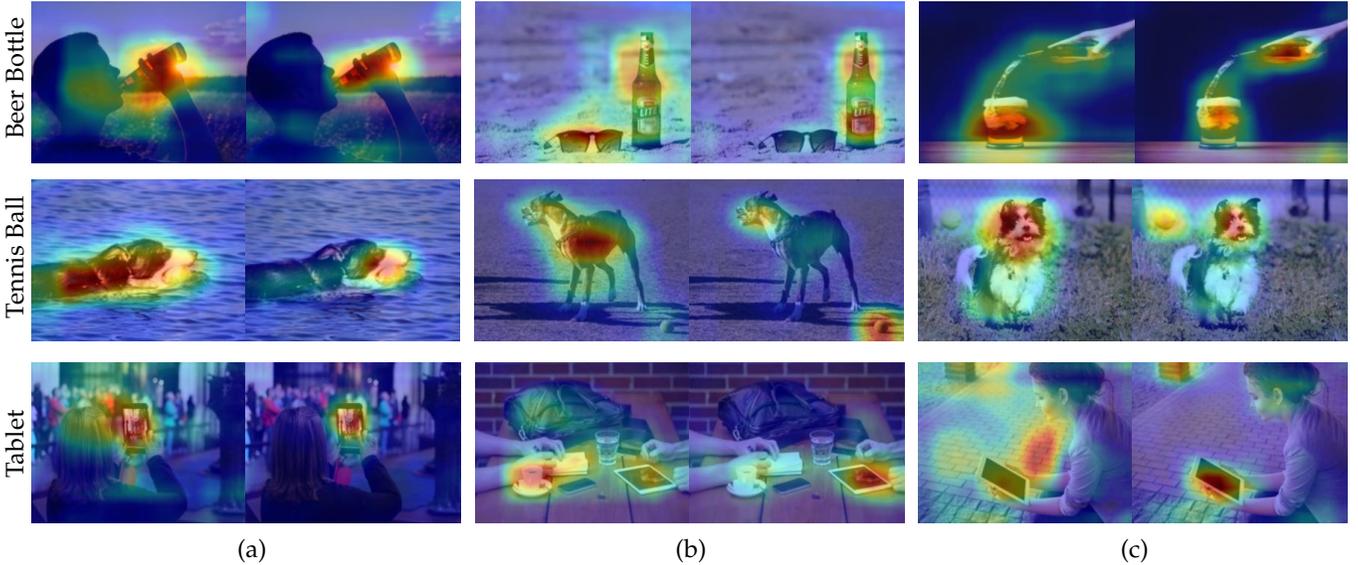

\textbf{Baseline.} 
We follow GCoNet~\cite{GCoNet} to design our \ourmodel{}~in a siamese way. Note that GCoNet follows the architecture of GICD~\cite{GICD} without extensive experiments on the validity of each component in GICD, including the multi-head supervision, loss function, feature normalization,~\etc{}. Although these components bring additional parameters and complexity to the network itself, experimental proof still does not support the effectiveness. Instead of taking these components as granted, we conduct extensive experiments on each component.
Firstly, we try to substitute the blocks of multiple convolutions in lateral connections with only one 1x1 convolution layer as the original FPN~\cite{FPN} does. Secondly, we try to remove the multi-stage supervision of the saliency maps on the decoder. Thirdly, we try to add batch normalization behind every convolution layer except 1x1 convolution layers. Finally, as our experiments show, BCE loss brings higher accuracy to our experiments, while IoU loss brings more binarized final saliency maps and faster convergence. To better combine the two losses, we control the initial BCE and IoU loss on the same quantity level with different weights and sum them up.

These modifications can be summarized into three parts, \ie{}, network architecture simplification, normalization layers, and the hybrid loss. 
Following Occam's Razor\footnote{\url{https://en.wikipedia.org/wiki/Occams\_razor}}, 
we try to remove all uncertain modules used in many existing works without enough experimental proof. These modifications improve our \ourmodel{}~in terms of both simplicity and accuracy with a large margin compared with the baseline model GCoNet (ID:1 in \tabref{table:ablation_modif_overall}). As shown in \tabref{table:ablation_modif_overall}, combining all of them yields 2.6\% and 2.8\% relative improvement on CoSOD3k and CoSal2015 in terms of E-measure, respectively. It also achieves 2.5\% E-measure relative improvement on CoCA, the most challenging CoSOD test set.

\textbf{Effectiveness of RACM.}
The RACM guides the model to learn more discriminative features to distinguish objects of different classes. Compared with the original ACM, it works more accurately and accelerates (see \figref{fig:racm_acm_lossTR}) the convergence of our \ourmodel{}. As seen in \tabref{table:ablation_modules}, the RACM slightly improves the baseline performance on CoCA and CoSOD3k in terms of most metrics. The activation maps in \figref{fig:CAM} show that our \ourmodel{}~gives a higher accuracy in various cases and guides the model to focus on the targets more precisely. The feature maps on each stage of the decoder in \ourmodel{}~are shown in \figref{fig:fpn_features}. As the results show, \ourmodel{}~has a better performance than GCoNet~\cite{GCoNet} in discriminating objects of different classes.

\begin{table*}[t!]
\begin{center}
\footnotesize
\renewcommand{\arraystretch}{1.0}
\renewcommand{\tabcolsep}{1.14mm}
\caption{\textbf{Quantitative comparisons between our \ourmodel{}~and other methods.} ``$\uparrow$'' (``$\downarrow$'') means that the higher (lower) is better. Methods are attached with links to open-source codes or paper sources. Since there are several datasets used in the CoSOD task for training, we list all the training sets used in corresponding methods, \ie{}, Train-1, 2, and 3 represent the DUTS\_class~\cite{GICD}, COCO-9k~\cite{GWD}, and COCO-SEG~\cite{COCO_SEG}, respectively.} 
\label{table:ablation_results}
\begin{tabular}{r||r|r|cccc|cccc|cccc}
\hline
 &  &  & \multicolumn{4}{c|}{CoCA~\cite{GICD}} & \multicolumn{4}{c|}{CoSOD3k~\cite{CoSOD3k}} & \multicolumn{4}{c}{CoSal2015~\cite{CoSal2015}} \\
Method & Pub. \& Year & Training Set & $E_{\xi}^\text{max} \uparrow$ & $S_\alpha \uparrow$ & $F_\beta^\text{ max} \uparrow$ & $\epsilon \downarrow$ & $E_{\xi}^\text{ max} \uparrow$ & $S_\alpha \uparrow$ & $F_\beta^\text{ max} \uparrow$ & $\epsilon \downarrow$ & $E_{\xi}^\text{ max} \uparrow$ & $S_\alpha \uparrow$ & $F_\beta^\text{ max} \uparrow$ & $\epsilon \downarrow$ \\
\hline
\href{https://github.com/HzFu/CoSaliency_tip2013}{CBCS}~\cite{CBCS} & TIP 2013 & - & 0.641 & 0.523 & 0.313 & 0.180 & 0.637 & 0.528 & 0.466 & 0.228 & 0.656 & 0.544 & 0.532 & 0.233 \\
\href{https://www.ijcai.org/proceedings/2017/0424.pdf}{GWD}~\cite{GWD} & IJCAI 2017 & Train-2 & 0.701 & 0.602 & 0.408 & 0.166 & 0.777 & 0.716 & 0.649 & 0.147 & 0.802 & 0.744 & 0.706 & 0.148 \\
\href{https://www.ijcai.org/proceedings/2019/0115.pdf}{RCAN}~\cite{RCAN} & IJCAI 2019 & Train-2 & 0.702 & 0.616 & 0.422 & 0.160 & 0.808 & 0.744 & 0.688 & 0.130 & 0.842 & 0.779 & 0.764 & 0.126 \\
\href{https://github.com/ltp1995/GCAGC-CVPR2020}{GCAGC}~\cite{GCAGC} & CVPR 2020 & Train-3 & 0.754 & 0.669 & 0.523 & 0.111 & 0.816 & 0.785 & 0.740 & 0.100 & 0.866 & 0.817 & 0.813 & 0.085 \\
\href{https://github.com/zzhanghub/gicd}{GICD}~\cite{GICD} & ECCV 2020 & Train-1 & 0.715 & 0.658 & 0.513 & 0.126 & 0.848 & 0.797 & 0.770 & 0.079 & 0.887 & 0.844 & 0.844 & 0.071 \\
\href{https://github.com/blanclist/ICNet}{ICNet}~\cite{ICNet} & NeurIPS 2020 & Train-2 & 0.698 & 0.651 & 0.506 & 0.148 & 0.832 & 0.780 & 0.743 & 0.097 & 0.900 & 0.856 & 0.855 & \textbf{0.058} \\
\href{https://github.com/KeeganZQJ/CoSOD-CoADNet}{CoADNet}~\cite{CoADNet} & NeurIPS 2020 & Train-1, 3 & - & - & - & - & 0.878 & 0.824 & 0.791 & 0.076 & 0.914 & 0.861 & 0.858 & 0.064 \\
\href{https://github.com/DengPingFan/CoEGNet}{CoEGNet}~\cite{CoSOD3k} & TPAMI 2021 & Train-1 & 0.717 & 0.612 & 0.493 & 0.106 & 0.837 & 0.778 & 0.758 & 0.084 & 0.884 & 0.838 & 0.836 & 0.078 \\
\href{https://openaccess.thecvf.com/content/CVPR2021/papers/Zhang_DeepACG_Co-Saliency_Detection_via_Semantic-Aware_Contrast_Gromov-Wasserstein_Distance_CVPR_2021_paper.pdf}{DeepACG}~\cite{DeepACG} & CVPR 2021 & Train-3 & 0.771 & 0.688 & 0.552 & 0.102 & 0.838 & 0.792 & 0.756 & 0.089 & 0.892 & 0.854 & 0.842 & 0.064 \\
\href{https://github.com/nnizhang/CADC}{CADC}~\cite{CADC} & ICCV 2021 & Train-1, 2 & 0.744 & 0.681 & 0.548 & 0.132 & 0.840 & 0.801 & 0.859 & 0.096 & 0.906 & 0.866 & 0.862 & 0.064 \\ 
\href{https://github.com/siyueyu/DCFM}{DCFM}~\cite{DCFM} & CVPR 2022 & Train-2 & 0.783 & 0.710 & 0.598 & 0.085 & 0.874 & 0.810 & 0.805 & 0.067 & 0.892 & 0.838 & 0.856 & 0.067 \\ 
\href{https://github.com/suyukun666/UFO}{UFO}~\cite{arxiv_UFO} & ArXiv 2022 & Train-3 & 0.782 & 0.697 & 0.571 & 0.095 & 0.874 & 0.819 & 0.797 & 0.073 & 0.906 & 0.860 & 0.865 & 0.064 \\ 
\hline

\textbf{\href{https://github.com/fanq15/GCoNet}{GCoNet}~(Ours)} & CVPR 2021 & Train-1 & 0.760 & 0.673 & 0.544 & 0.105 & 0.860 & 0.802 & 0.777 & 0.071 & 0.887 & 0.845 & 0.847 & 0.068 \\
\textbf{\href{https://github.com/ZhengPeng7/GCoNet_plus}{\ourmodel{}}~(Ours)}	& Submission & Train-1 & 0.786 & 0.691 & 0.574 & 0.113 & 0.881 & 0.828 & 0.807 & 0.068 & 0.917 & 0.875 & 0.876 & 0.054 \\
\textbf{\href{https://github.com/ZhengPeng7/GCoNet_plus}{\ourmodel{}}~(Ours)}	& Submission & Train-2 & 0.798 & 0.717 & 0.605 & 0.098 & 0.877 & 0.819 & 0.796 & 0.075 & 0.902 & 0.853 & 0.857 & 0.073 \\
\textbf{\href{https://github.com/ZhengPeng7/GCoNet_plus}{\ourmodel{}}~(Ours)}	& Submission & Train-3 & 0.787 & 0.712 & 0.602 & 0.100 & 0.875 & 0.820 & 0.793 & 0.075 & 0.899 & 0.853 & 0.852 & 0.071 \\
\textbf{\href{https://github.com/ZhengPeng7/GCoNet_plus}{\ourmodel{}}~(Ours)}	& Submission & Train-1, 2 & 0.808 & 0.734 & 0.626 & 0.088 & 0.894 & 0.839 & 0.822 & 0.065 & 0.919 & 0.876 & 0.880 & 0.058 \\
\rowcolor{mygray}
\textbf{\href{https://github.com/ZhengPeng7/GCoNet_plus}{\ourmodel{}}~(Ours)}	& Submission & Train-1, 3 & \textbf{0.814} & \textbf{0.738} & \textbf{0.637} & \textbf{0.081} & \textbf{0.901} & \textbf{0.843} & \textbf{0.834} & \textbf{0.062} & \textbf{0.924} & \textbf{0.881} & \textbf{0.891} & \textbf{0.056} \\
\hline
\end{tabular}
\label{table:main}
\end{center}
\end{table*}

\textbf{Effectiveness of CEM.} 
Among the previous CoSOD methods, IoU and BCE loss tend to be employed as the training loss. However, in most of these methods, only one single loss is used for the supervision during training. BCE guides the supervision from the pixel perspective, and IoU guides the supervision from the view of regions. Despite the outstanding performance achieved by many existing methods~\cite{GCAGC,CoADNet,DeepACG,GCoNet,DCFM}, using the BCE and IoU separately suffers from some issues. Specifically, with IoU loss supervising the model on the region level, the predicted saliency maps are usually rough and cannot handle the small details very well. BCE can guide the model to focus on the details. At the same time, saliency maps supervised with it tend to contain much uncertainty, which makes the predictions challenging to use in the application directly.
In this case, we apply the CEM to simultaneously predict more accurate and binarized maps closer to the demand of real-world applications. As shown in \figref{fig:qualitive_ablation} and \tabref{table:ablation_modules}, the CEM can make the predicted maps better in terms of both accuracy and visualization.

\textbf{Effectiveness of GST Loss.} 
Consensus features play an important role in the CoSOD task for detecting common objects. However, the consensus features of some categories may get too close to each other. To this end, we need to keep consensus features more distinguishing and the distance far away from other features. We introduce the GST loss to make features of different classes learned more discriminative to each other. As experiments show in \tabref{table:ablation_modules} and \figref{fig:qualitive_ablation}, GST loss successfully differentiates the features on a global and RoI level and further improves the model's competitiveness.

\subsection{Competing Methods}
Since not all CoSOD models are publicly available, we only compare our GCoNet and \ourmodel{}~with one representative traditional algorithm CBCS~\cite{CBCS} and \numDeepMethods deep-learning based CoSOD models, including all update-to-date models, \ie{}, GWD~\cite{GWD}, RCAN~\cite{RCAN}, CSMG~\cite{CSMG}, GCAGC~\cite{GCAGC}, GICD~\cite{GICD}, ICNet~\cite{ICNet}, CoADNet~\cite{CoADNet}, CoEGNet~\cite{CoSOD3k}, DeepACG~\cite{DeepACG}, CADC~\cite{CADC}, UFO~\cite{arxiv_UFO}, and DCFM~\cite{DCFM}. Because of the much more excellent performance of the latest CoSOD methods compared with single-SOD methods, we do not list the single-SOD ones. A complete leaderboard of previous methods can be found in~\cite{CoSOD3k}.

\begin{figure*}[t!]
	\centering
    \begin{overpic}[width=.96\textwidth]{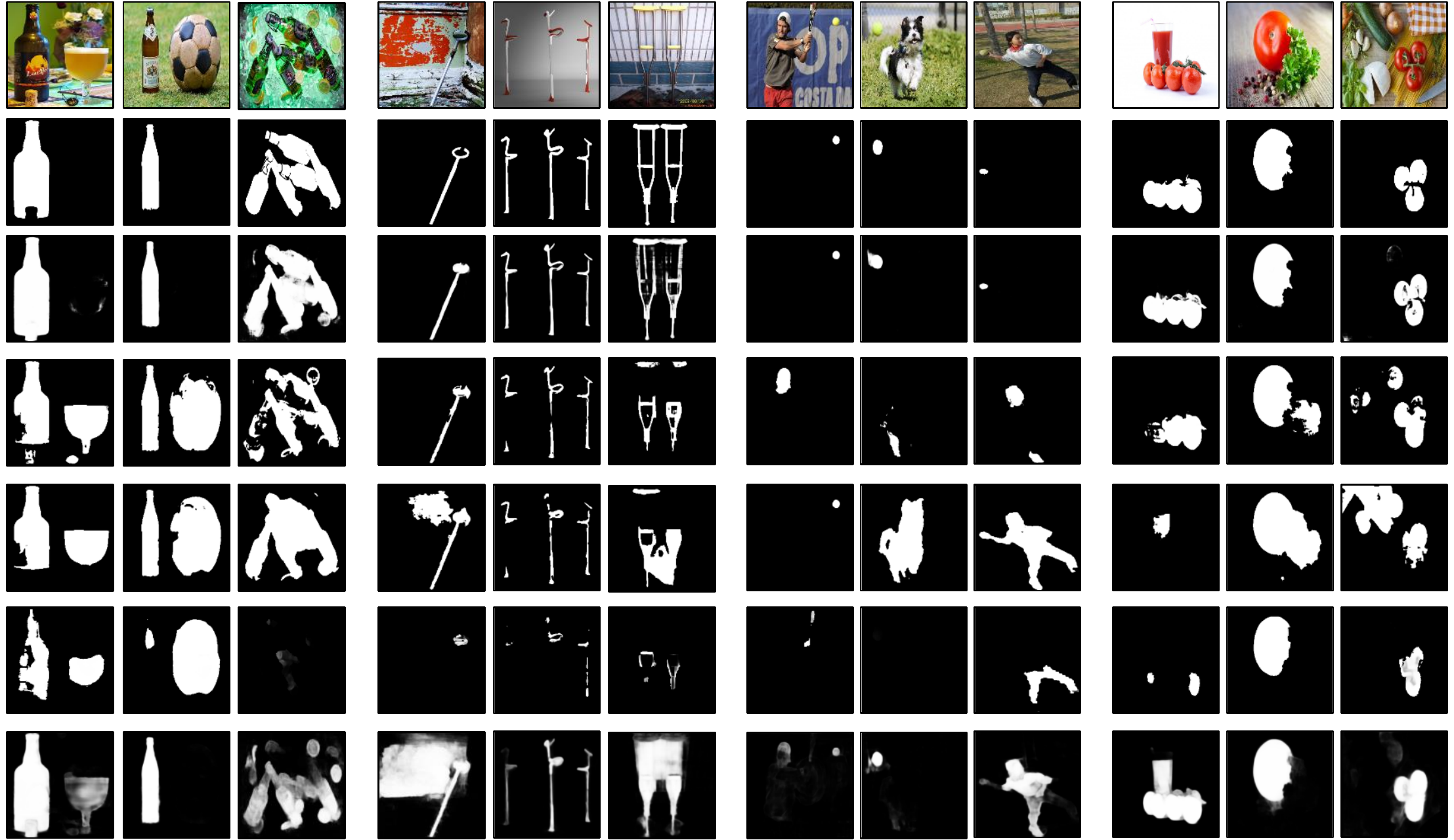}
        \put(-1.5,52){\rotatebox{90}{Input}}
        \put(-1.5,45){\rotatebox{90}{GT}}
        \put(-1.5,34){\rotatebox{90}{\textbf{\ourmodel{}}}}
        \put(-1.5,26){\rotatebox{90}{GCoNet}}
        \put(-1.5,18.5){\rotatebox{90}{GICD}}
        \put(-1.5,10){\rotatebox{90}{CoEG}}
        \put(-1.5,1){\rotatebox{90}{CADC}}
        \put(7,-2){{Beer Bottle}}
        \put(34.5,-2){{Crutch}}
        \put(60,-2){{Tennis}}
        \put(85,-2){{Tomato}}
    \end{overpic}
	\caption{\textbf{Qualitative comparisons of our \ourmodel{}~and other methods.} ``GT'' denotes the ground truth. The predictions in the row of \textbf{Ours} are produced by our \ourmodel{}~, which is trained with the DUTS\_class dataset.}
	\label{fig:qualitive_res}
\end{figure*}
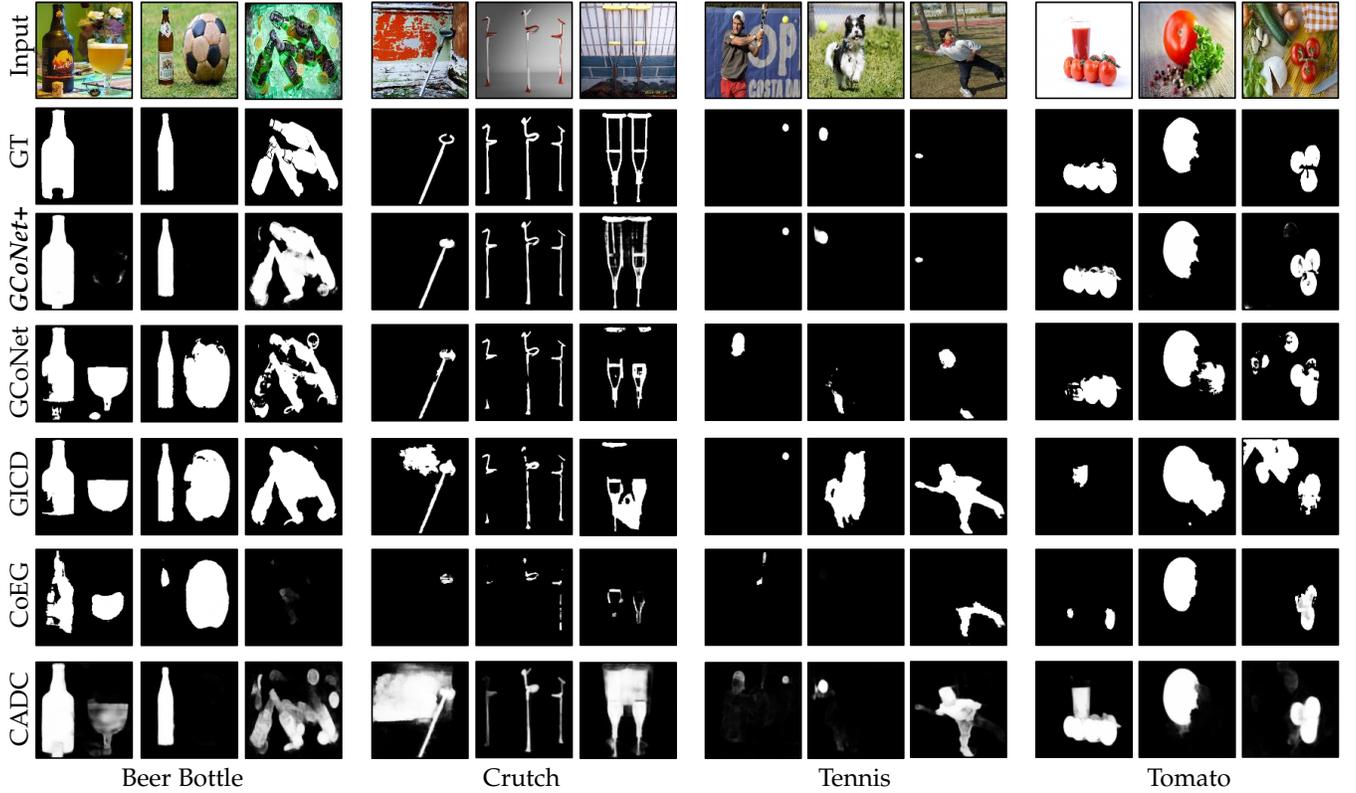

\textbf{Quantitative Results.}
\tabref{table:ablation_results} shows the quantitative results of our \ourmodel{}~and previous state-of-the-art methods. Our \ourmodel{}~outperforms all of them in all metrics, especially on the CoCA and CoSOD3k datasets. CoCA is the most difficult dataset to specify the common objects compared with two other datasets because of the larger number of objects in a single image and more diverse backgrounds. Our \ourmodel{}~shows a stronger ability in segmentation, which benefits from the improved features in terms of saliency detection and consensus learning, respectively. CoSOD3k has similar attributes, and our \ourmodel{}~keeps its best performance over all other methods on this dataset. CoSal2015 is the easiest dataset since most of its images contain only one salient object, making it easy to handle with single-SOD methods. Despite the low difficulty and absence of co-saliency, our \ourmodel{}~still outperforms other methods with a relatively smaller margin. Besides, our \ourmodel{}~ has fewer parameters and makes the faster inference compared with most of the existing methods, as shown in \tabref{table:ablation_fps}.

\begin{table}[t!]
\begin{center}
\footnotesize
\renewcommand{\arraystretch}{1.0}
\renewcommand{\tabcolsep}{9.0pt}
\caption{\textbf{Runtime comparisons of different methods.} Batch size is set as 2 for all methods during their inference on an A100 GPU.}
\label{table:ablation_fps}
\vspace{-10pt}
\begin{tabular}{r|c|c}
\hline
    Methods     & Inference Time (ms)    & Parameters (MB)    \\
    \hline
    {GICD}~\cite{GICD}        & 7.1       & 1060.7      \\
    {ICNet}~\cite{ICNet}      & 6.6       & \textbf{70.3}      \\
    {CoADNet}~\cite{CoADNet}     & 13.1       & 113.2      \\
    {GCAGC}~\cite{GCAGC}         & 58.1       & 280.7      \\
    {CADC}~\cite{CADC}     & 58.0      & 1498.7      \\
    {DCFM}~\cite{DCFM}     & 4.6      & 542.9      \\
    \hline
    {GCoNet}~\cite{GCoNet} (\textbf{Ours})    & \textbf{2.1}      & 541.7      \\
    \rowcolor{mygray}
    \textbf{\ourmodel{}~(Ours)}        & 3.5       & \textbf{70.3}      \\
    \hline
\end{tabular}
\end{center}
\end{table}

\textbf{Qualitative Results.}
\figref{fig:qualitive_res} shows the saliency maps generated by different methods for qualitative comparison. Images of beer\_bottle group contain multiple salient objects of numerous classes, where our \ourmodel{}~can precisely detect the co-salient objects while others cannot. In the crutch group, targets are slim sketches, but our \ourmodel{}~can still segment the sketches with high accuracy while others even fail to make the correct segmentation. We put the tennis group here to compare the ability of models to detect small objects, where our \ourmodel{}~performs better than others in terms of both classification and precision. In contrast, others may miss the small objects or focus on objects with other classes. Many tomatoes appear as co-salient objects in the tomato group and should be detected simultaneously. Our method can find all the tomatoes with a good saliency map, but others may omit some of these salient tomatoes or segment other very close objects. Among the examples above, our \ourmodel{}~better finds the intra-group common information and discriminates the inter-group information. 

\begin{figure}[t!]
	\centering
    \begin{overpic}[width=\linewidth]{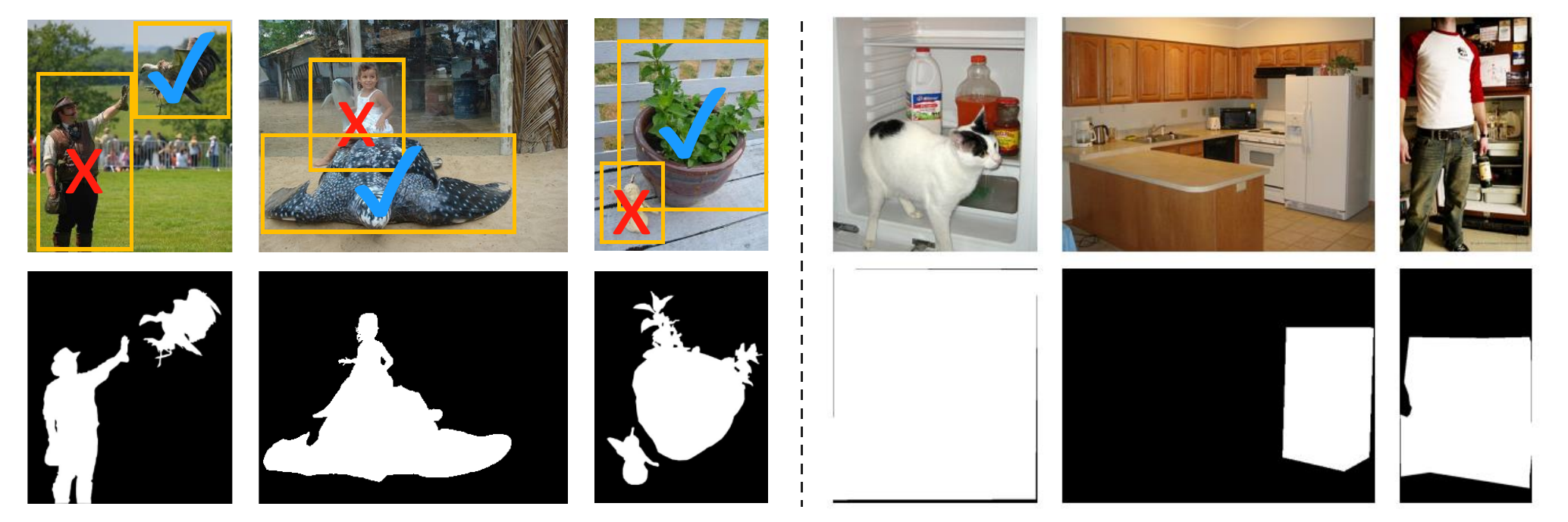}
        \put(16,-3){{DUTS\_class}}
        \put(53,-3){{COCO-9k and COCO-SEG}}
    \end{overpic}
	\caption{\textbf{Ground truth with problem occurs in DUTS\_class dataset, COCO-9k, and COCO-SEG dataset.} As the examples given show, salient objects of different classes occur together in one image of DUTS\_class dataset, whose regions are wrong ground truth in DUTS\_class. In COCO-9k and COCO-SEG, many objects are not salient, which also plays a bad role in training a CoSOD model.}
	\label{fig:duts_coco}
\end{figure}

\subsection{Discussion of Existing CoSOD Training Sets}
\label{sec:training_set_problem}
Even though many great works have been proposed in CoSOD, there remains a lack of a standard training set. DUTS\_class, COCO-9k, and COCO-SEG are three commonly-used training sets but have their limitations,~\eg{}, inaccurate GT maps, and a small number of target objects.

\begin{figure*}[t!]
	\centering
    \begin{overpic}[width=.96\textwidth]{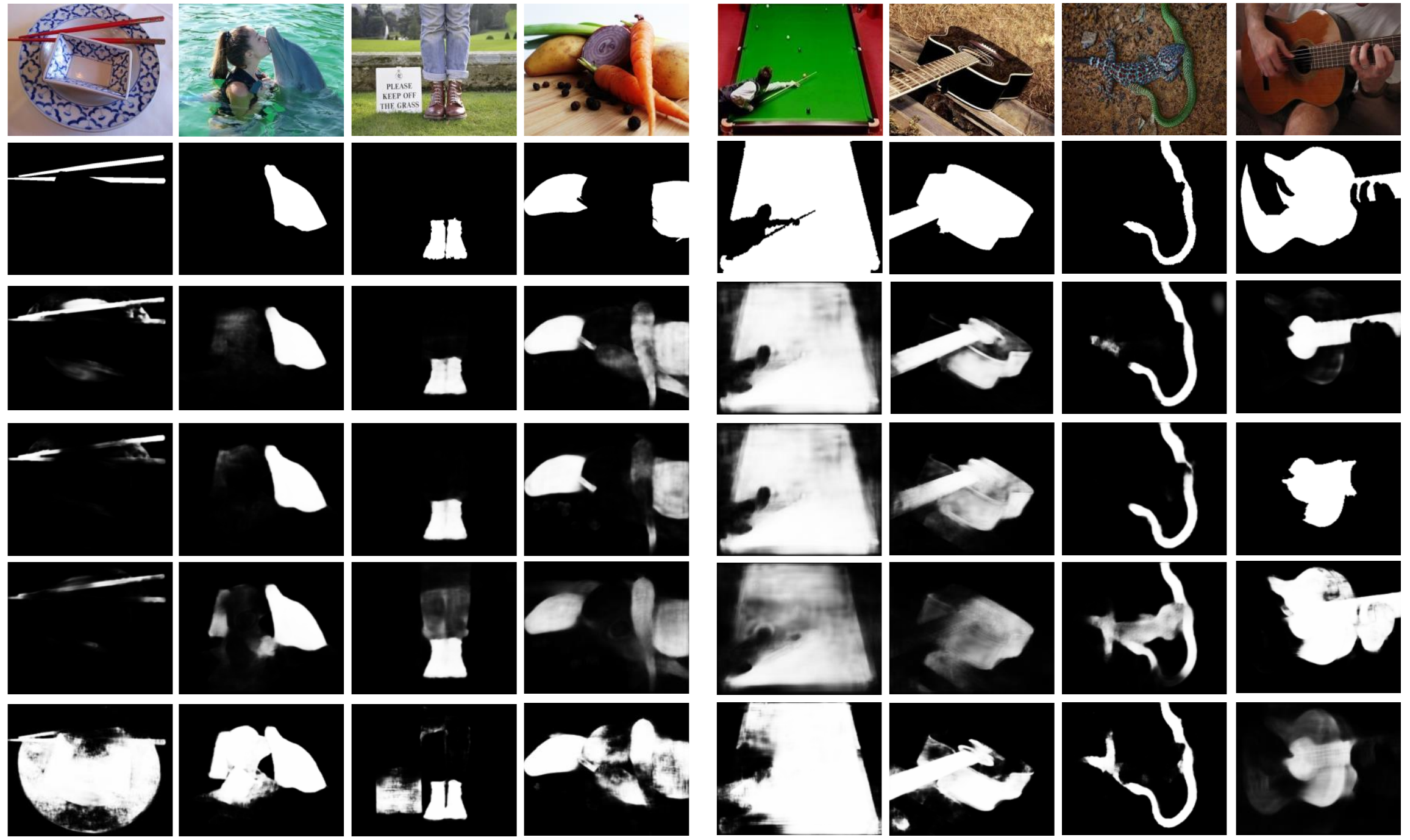}
        \put(-1.5,52.5){\rotatebox{90}{Input}}
        \put(-1.5,43.5){\rotatebox{90}{GT}}
        \put(-1.5,31.5){\rotatebox{90}{Mix-1, 3}}
        \put(-1.5,21.5){\rotatebox{90}{Mix-1, 2}}
        \put(-1.5,11.0){\rotatebox{90}{COCO-9k}}
        \put(-1.5,0.1){\rotatebox{90}{DUTS\_class}}
        \put(22,-2){{CoCA}}
        \put(71,-2){{CoSal2015}}
    \end{overpic}
    \vspace{0mm}
	\caption{
        \textbf{Qualitative results produced by \ourmodel{}~ trained on different training sets.} We adopt different datasets to set up experiments to validate the different optimization directions of the DUTS\_class dataset~\cite{DUTS} and COCO-9k/COCO-SEG datasets~\cite{GWD,COCO_SEG}. ``Mix-1, 2'' denotes both the DUTS\_class and COCO-9k are used in training, ``Mix-1, 3'' denotes both the DUTS\_class and COCO-SEG are used in training. CoCA is the most complex CoSOD test set and needs more attention to be paid to finding objects of the common class. At the same time, CoSal2015 is a relatively simple one that almost only measures the ability of salient object detection in most cases.
	}\label{fig:qualitive_res_duts_coco9k}
\end{figure*}
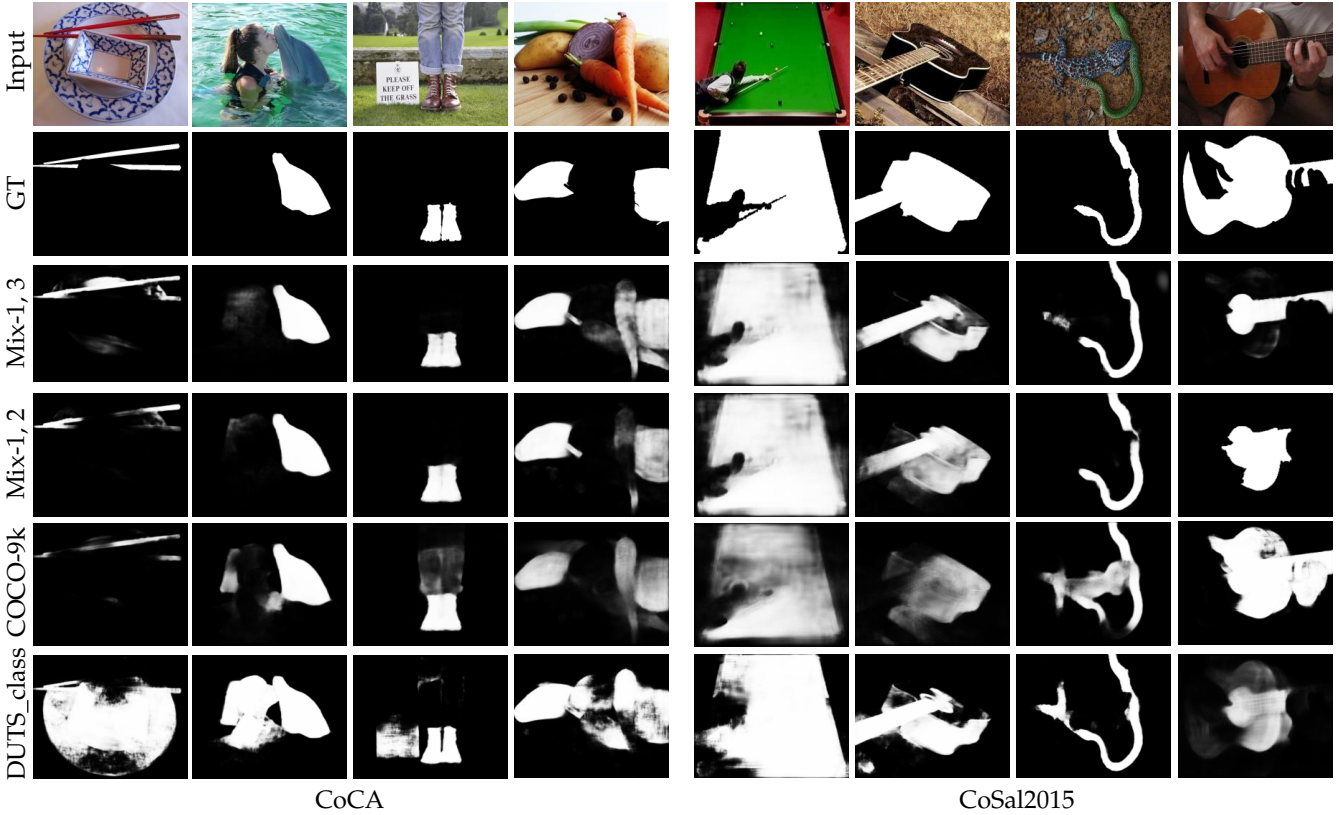

\textbf{DUTS\_class.}
Due to DUTS\_class aiming only at detecting salient objects, there are salient objects of different classes in a single image. As \figref{fig:duts_coco} shows, objects with wrong categories still exist in the ground truth, which gives the models a wrong optimization direction. Whatsmore, there are only a very few target objects in a single image, which makes the training lack the segmentation ability of common objects.

\textbf{COCO-9k/COCO-SEG.}
As mentioned in~\cite{GWD,COCO_SEG}, COCO-9k and COCO-SEG are both collected from COCO dataset~\cite{COCO}. However, neither of them takes the salient objects into account. Therefore, objects with ground truth may not be salient. Thus, models trained on only COCO-9k or only COCO-SEG may perform well on segmenting common objects while badly on segmenting salient objects.

\textbf{Experiments.}
Among the three public test sets and realistic scenarios, cases can be difficult or easy, with various objects and complex contexts, or simply a dominant object on a white paper. To deal with all these cases with a satisfying result, the models need to behave well on both the common object segmentation and salient object detection, which are the main optimization goals that can be learned from COCO-9k~\cite{GWD}/COCO-SEG~\cite{COCO_SEG} and DUTS\_class~\cite{GICD}, respectively. As mentioned in \secref{sec:datasets}, CoCA~\cite{GICD} focuses more on segmenting the common objects in complex contexts, while CoSal2015~\cite{CoSal2015} plays a more critical role in testing the ability of models to detect salient objects. 
We take these two datasets to check the different aspects of the model's performance.

We train \ourmodel{}~on the DUTS\_class and COCO-9k/COCO-SEG both separately and jointly. Taking the results of CoSal2015~\cite{CoSal2015} shown in \figref{fig:qualitive_res_duts_coco9k}, models trained on DUTS\_class~\cite{GICD} show a better performance on SOD tasks but weakness on detecting objects of the common class. 
However, getting trained on COCO-9k or COCO-SEG enables the model to learn a good ability to segment objects with a common class while having a relatively worse performance on detecting simple salient objects. Compared with models training on DUTS\_class, models trained only on COCO-9k/COCO-9k often fail to detect the salient objects.

To deal with the two sub-tasks in CoSOD, \ie{}, segmenting common objects and detecting salient objects, we need to optimize our \ourmodel{}~in two directions. Therefore, we set a joint training of our \ourmodel{}~on DUTS\_class~\cite{GICD} and COCO-9k/COCO-SEG~\cite{GWD,COCO_SEG}. Under the setting of combined training, the same model shows more robust performance in both the two directions mentioned above. As the performance shown in  \tabref{table:ablation_results}, the jointly trained (\ie{}, Train-1, 3) model achieves much better results on all these three test sets. Specifically, compared with the model trained only on DUTS\_class, our \ourmodel{}~shows comparable performance on CoSal2015 while much better performance on CoCA. Meanwhile, compared with the model trained only COCO-9k or only COCO-SEG, the jointly trained model shows similar performance on CoCA while much better results on CoSal2015. The same phenomenon also occurs on the predicted maps, as shown in \figref{fig:qualitive_res_duts_coco9k}.

\subsection{Failure Cases}
\label{sec:fail}

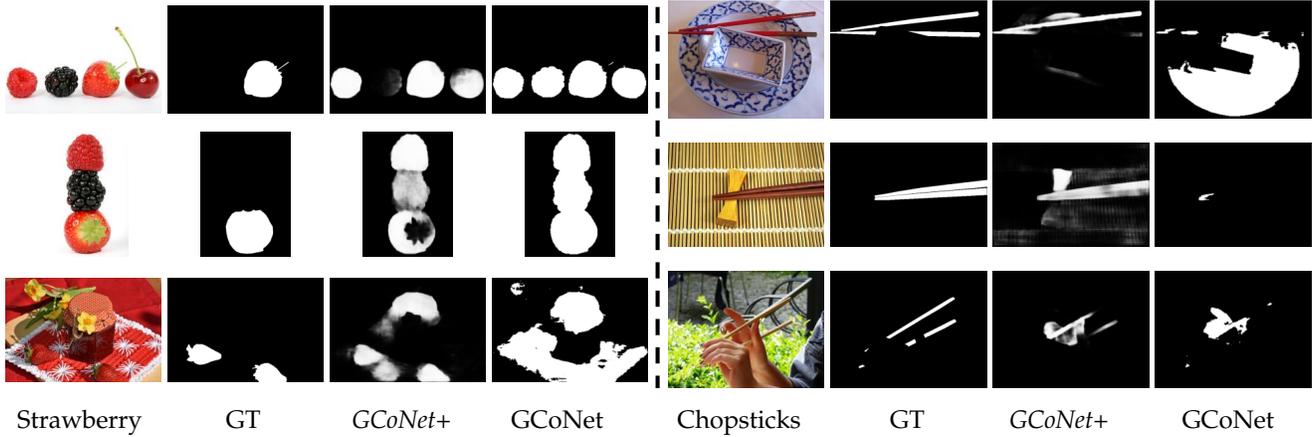
\begin{figure*}[t!]
	\centering
    \begin{overpic}[width=.97\textwidth]{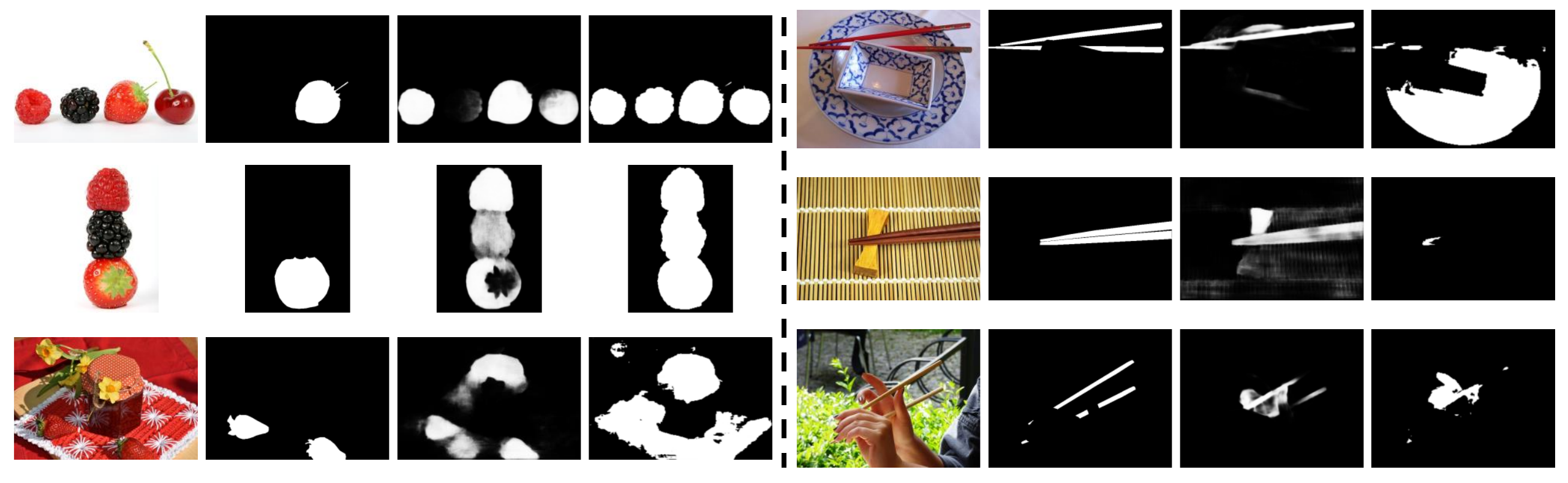}
        \put(1.8,-2){{Strawberry}}
        \put(17.5,-2){{GT}}
        \put(27,-2){{\ourmodel{}}}
        \put(39,-2){{GCoNet}}
        \put(51.5,-2){{Chopsticks}}
        \put(67.5,-2){{GT}}
        \put(76.5,-2){{\ourmodel{}}}
        \put(89.5,-2){{GCoNet}}
    \end{overpic}
    \vspace{2mm}
	\caption{
    \textbf{Failure cases in the results produced by our \ourmodel{}.} We provide the typical failure cases made by our \ourmodel{} and GCoNet~\cite{GCoNet}. On the left side, the inaccurate classification of the surrounding objects makes the bad CoSOD results. On the right side, there lacks the ability of fine segmentation making inaccurate predictions of saliency maps.
	}\label{fig:failures}
\end{figure*}

There are two main sub-goals of the co-salient object detection networks, which means we can delineate the ability of the network from two perspectives, \ie{}, finding the common objects and segmenting the salient ones of them. Thus, we select the typical failures of these two types for analysis. As shown in \figref{fig:failures}, when too many similar objects of different classes appear in a single image or target objects are hard to separate from neighboring objects, our models may misidentify them and give inaccurate predictions.

Specifically, for the strawberries shown on the left of \figref{fig:failures}, our well-trained \ourmodel{} tends to focus on the texture and color of objects. The pockmark textures can be  misidentified as those of strawberries. Thus, the cranberries and cherries are misidentified as strawberries while the blueberries can be identified. For the chopsticks on the right side in the figure, our \ourmodel{} is more capable of finding the target objects while still cannot handle the difficult segmentation problem.
Although our \ourmodel{} still faces some problems in these very difficult cases, it still demonstrates high potential while outperforming our previous GCoNet.

To further improve the models in these difficult cases, a larger training set consisting of more classes will be needed. A larger number of classes will bring forth a stronger ability to discriminate similar objects of different classes, while more segmentation examples will enhance the general segmentation ability to accurately segment objects in complex scenarios. As mentioned in \secref{sec:training_set_problem}, this may be a major potential contribution to the CoSOD task in the future.


\section{Potential Applications}
We show the potential to utilize the extracted co-saliency maps to produce segmentation masks of high quality for related downstream image processing tasks.

\begin{figure}[t!]
\begin{center}
\includegraphics[width=.98\linewidth]{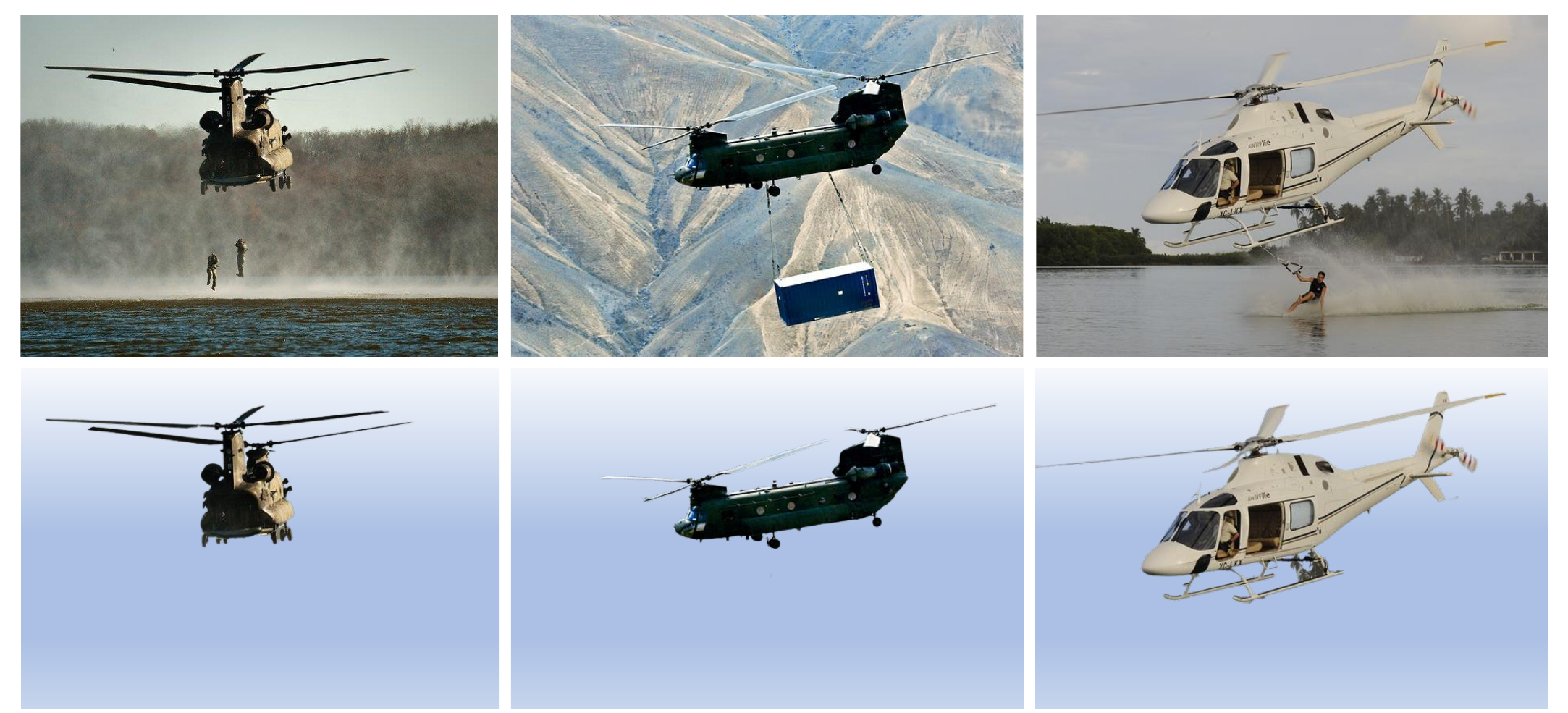}
\end{center}
\vspace{-10pt}
\caption{\textbf{Application \#1.} Content aware object co-segmentation visual results (``Helicopter'') obtained by our \ourmodel{}.}
\label{fig:app1}
\end{figure}

\textbf{Application \#1: Content-Aware Co-Segmentation.}
Co-saliency maps have been widely used in image pre-processing tasks. Taking the unsupervised object segmentation in our implementation as an example, we first find a group of images by keywords on the Internet. Then, our \ourmodel{}~is applied to generate co-saliency maps. Finally, the salient objects of the specific group can be extracted with the co-saliency maps. Following~\cite{Cheng2011GlobalCB}, we can use GrabCut~\cite{grabcut} to obtain the final segmentation results. Adaptive threshold~\cite{Peng2014RGBDSO} is chosen here to initialize GrabCut for the binary version of the saliency maps. As shown in \figref{fig:app1}, our method works well in the content-aware object co-segmentation task, which should benefit existing E-commerce applications in the background replacement.

\begin{figure}[t!]
\begin{center}
\includegraphics[width=.98\linewidth]{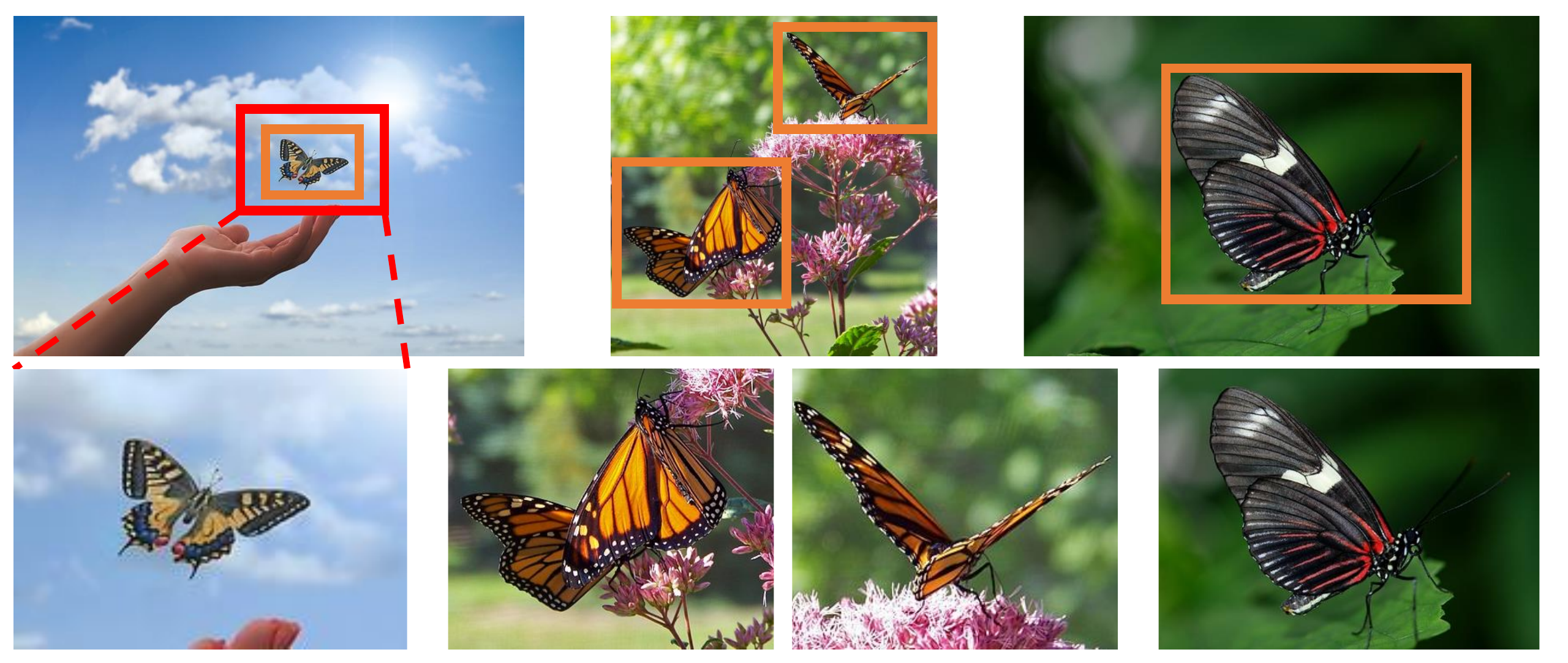}
\end{center}
\vspace{-10pt}
\caption{\textbf{Application \#2.} Co-localization based automatic thumbnails (``Butterfly'') generated by our \ourmodel{}.}
\label{fig:app2}
\end{figure}

\textbf{Application \#2: Automatic Thumbnails.}
The idea of paired-image thumbnails is derived from~\cite{CoSOD_trad_align}. With the same goal\footnote{Jacobs~\etal{}’s work~\cite{CoSOD_trad_align} is limited to the case of image pairs}, we introduce a CNN-based application of photographic triage, which is valuable for sharing images on the website. As \figref{fig:app2} shows, the orange box is generated by the saliency maps obtained from \ourmodel{}. We can also scale them up with the orange box and get the larger red one. Finally, the collection-aware crops technique~\cite{CoSOD_trad_align} can be adapted to produce the results shown in the second row.

\section{Conclusion}
This work proposes a novel group collaborative model (\ourmodel{}) to deal with the CoSOD task. With the experiments conducted, we find that group-level consensus can introduce effective semantic information, auxiliary classification, and metric learning to improve the feature representation in terms of intra-group compactness  and inter-group separability. Qualitative and quantitative experiments demonstrate the superiority and state-of-the-art performance of our \ourmodel{}. 
We show that the techniques of \ourmodel{}~can also be transferred and easily used in many relevant applications, such as co-detection and co-segmentation.


\ifCLASSOPTIONcompsoc
  \section*{Acknowledgments}
\else
  \section*{Acknowledgment}
\fi
The authors would like to thank the anonymous reviewers and editor for their helpful comments on this manuscript. We thank Prof. Ling Shao for his insightful feedback. This work is partially supported by Huazhu Fu’s A*STAR Central Research Fund, and Career Development Fund (C222812010). This work is also supported by the National Natural Science Foundation of China (No. 62276129).

{
\bibliographystyle{IEEEtran}
\bibliography{TPAMI2021_GCoNet,ref_response}
}

\vfill

\end{document}